\documentclass[runningheads]{llncs}

\usepackage{eccv}

\usepackage{eccvabbrv}

\usepackage{graphicx}
\usepackage{booktabs}

\usepackage[accsupp]{axessibility}  % Improves PDF readability for those with disabilities.

\usepackage{hyperref}

\usepackage{orcidlink}

\usepackage{amsmath} 
\usepackage{amssymb} 
\usepackage{booktabs}
\usepackage{algorithmic}
\usepackage{setspace}
\usepackage{graphicx}
\usepackage{textcomp}
\usepackage[table]{xcolor}
\usepackage{multirow}
\usepackage{threeparttable}
\usepackage{tabularx}
\usepackage{mathtools}
\usepackage{float}
\newcommand{\rom}[1]{\uppercase\expandafter{\romannumeral #1\relax}}
\usepackage{pgfplots}
\pgfplotsset{compat=1.18}
\usepackage{pgfplotstable}
\usepackage{array}
\definecolor{ourgray}{gray}{0.9}
\usepackage{arydshln}
\usepackage{makecell}
\usepackage{enumitem} 
\usepackage{wrapfig}

\usepackage{siunitx}
\newcommand{\tablesize}{\fontsize{7pt}{8.0pt}\selectfont}
\newcommand{\tinysize}{\fontsize{6pt}{7.2pt}\selectfont}
\newcommand{\hdr}[2][0.9]{\textbf{\scalebox{#1}[1]{#2}}}

\makeatletter
\newcommand\blfootnote[1]{%
  \begingroup
  \renewcommand\thefootnote{}%
  \renewcommand\@makefnmark{}%
  \footnotetext{#1}%
  \addtocounter{footnote}{-1}%
  \endgroup
}
\makeatother

\newcommand{\cmirule}{\noindent\rule{\linewidth}{0.4pt}}

\newcommand{\tocitem}[3]{\noindent\textbf{#1}\hspace{0.6em}#2\dotfill #3\par\vspace{2pt}}
\newcommand{\tocsubitem}[3]{\noindent\hspace{1.6em}\textbf{#1}\hspace{0.6em}#2\dotfill #3\par\vspace{1pt}}

\begin{document}

\title{OP3DSG: Open-Vocabulary Part-Aware 3D Scene Graph Generation for Real-World Environments} 

\titlerunning{OP3DSG}

\author{
Yirum Kim\inst{1}\orcidlink{0009-0000-4494-1839} \and
Ue-Hwan Kim\inst{1, \dagger}\orcidlink{0000-0003-2201-2988}
% Ue-Hwan Kim\inst{1}\thanks{\scriptsize Corresponding author.}\orcidlink{0000-0003-2201-2988}
}

\authorrunning{Y. Kim and UH. Kim}

\institute{Gwangju Institute of Science and Technology (GIST), Gwangju, Republic of Korea\\
\email{kimyirum@gm.gist.ac.kr, uehwan@gist.ac.kr}}

\maketitle
\blfootnote{\scriptsize $\dagger$ Corresponding author}
\blfootnote{\scriptsize Project page: \url{https://k2room.github.io/OP3DSG}}

\begin{figure*}[h]
    \vspace{-7mm}
    \centering
    \includegraphics[width=1\textwidth]{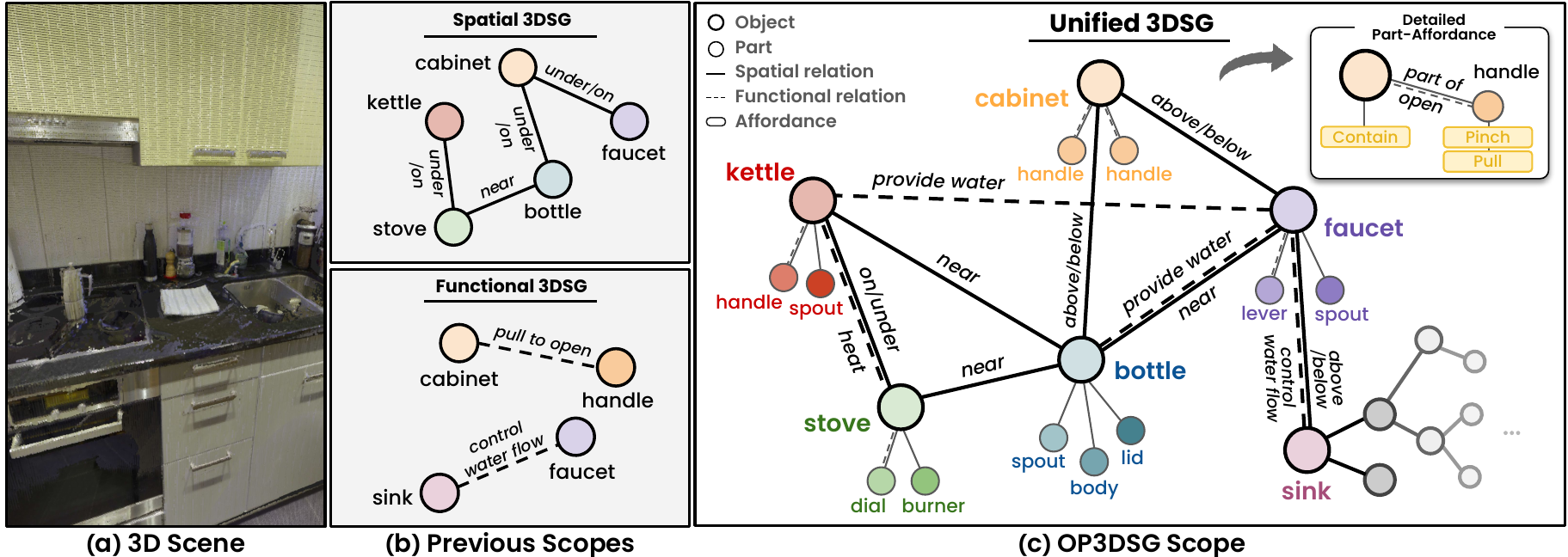}
    \caption{\textbf{Conceptual Comparison of 3D Scene Graphs (3DSGs).} (a) Given a 3D scene, (b) prior work falls into two scopes: \textit{Spatial 3DSGs}, which recognize objects and encode spatial relations, and \textit{Functional 3DSGs}, which mainly recognize interactive parts and functional relations. (c) Our scope unifies these notions into \textit{Unified 3DSGs}.}
    \label{fig:teasure}
    \vspace{-10mm}
\end{figure*}

\begin{abstract}
3D scene graphs (3DSGs) provide a compact and structured abstraction of 3D environments. Although advances in foundation models have enabled open-vocabulary 3DSG generation, existing approaches remain object-centric and encode limited relational information---restricting their applicability in real-world scenarios that require fine-grained understanding. We propose OP3DSG, an open-vocabulary part-aware 3DSG generation framework that constructs unified graphs that jointly model objects, interactive parts, spatial relations, functional relations, and affordances. OP3DSG integrates object-part knowledge-guided detection with part-aware 3D fusion to preserve small and interaction-relevant components, and employs a geometry-initialized prior graph with LLM-based refinement to reduce spurious relational predictions while enabling efficient graph construction. To systematically evaluate unified 3D scene graph construction, we introduce UniGraph3D, a benchmark designed for part-aware perception and multi-level relational reasoning. Experimental results show that OP3DSG achieves state-of-the-art performance and demonstrates its effectiveness as a perception backbone in diverse real-world robotics tasks. 
% Project page: \url{https://k2room.github.io/OP3DSG}
% See our \href{https://k2room.github.io/OP3DSG}{project page}.
  \keywords{3D Scene Graph \and Embodied AI \and Scene Understanding}
  % \keywords{3D Scene Graph \and Open-vocabulary Perception \and Embodied AI \and Scene Understanding}
\end{abstract}

\section{Introduction}
\label{sec:intro}

3D scene graphs (3DSGs) are a compact, structured abstraction that connects geometry with semantics and relationships, enabling efficient querying and reasoning over large environments~\cite{kim20193,wald2020learning,wu2021scenegraphfusion,wu2023incremental}.
Such structure is particularly attractive for embodied agents---conducting various downstream tasks such as navigation~\cite{li2022remote, gadre2022continuous} and task planning~\cite{Agia2022TaskographyER, Rana2023SayPlanGL}---where a graph can serve as a persistent, queryable ``mental model'' over long horizons and across viewpoints~\cite{gadre2022continuous,Agia2022TaskographyER}.
Recent progress has pushed 3DSGs toward \emph{open-vocabulary} settings by leveraging vision--language foundation models and open-set detectors~\cite{li2022grounded,zhou2022detecting,liu2024grounding}.
This enables graphs to be queried with novel category names beyond closed label spaces---leading to open-vocabulary 3DSG construction pipelines~\cite{koch2024open3dsg,gu2024conceptgraphs}.
In parallel, a complementary line of work argues that \emph{functionality} must be part of the representation---proposing functional 3DSGs that model interactive elements and functional relations~\cite{delitzas2024scenefun3d,zhang2025open}.

Despite these advances, current open-vocabulary 3DSGs remain largely \emph{object-centric} and often fail where physical interaction demands precision: small components (\eg, buttons, handles, switches) and fine-grained dependencies (\eg, a remote controlling a TV, a knob operating a burner) are missing or poorly grounded. On the other hand, the function-centric representation remains incomplete for general-purpose embodied reasoning: spatial structure, object-part hierarchy, and affordance are often handled in separate modules. Consequently, task planning that reasons over object-level graphs can navigate to a ``microwave'' but cannot decompose ``heat food'' into sub-goals like ``press start button,''  whereas reasoning over function-centric graphs does not generate high-level plans. Put simply, embodied agents require a \emph{compositional} view of the world: not only \textit{what} is present, but \textit{which part} is actionable and \textit{how} that part enables a function.

\begin{wrapfigure}[19]{r}{6.2cm}
    \centering
    \vspace{-7mm}
    \includegraphics[width=1\linewidth]{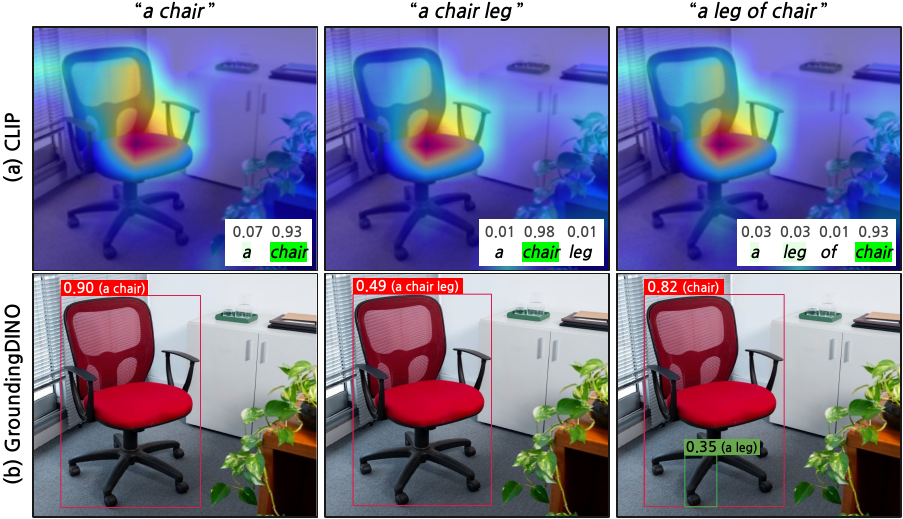}
    \caption{\textbf{Preliminary experiments reveal limitations in part detection.} (a) With CLIP~\cite{radford2021learning}, part-centric phrases (\eg, \textit{“a chair leg”, “a leg of chair”}) yield activation maps nearly identical to the object-level prompt \textit{“a chair”}, as attention concentrates on the token \textit{chair}. (b) With GroundingDINO~\cite{liu2024grounding}, part-centric phrases still yield object-level bounding boxes; when a part is detected, its confidence score is notably low.}
    \label{fig:intro}
\end{wrapfigure}
We take a step toward this goal by introducing the \textbf{Unified 3D Scene Graph} (\cref{fig:teasure}): a single graph that jointly models \emph{objects}, \emph{interactive parts}, \emph{functional relations}, \emph{spatial relations}, and \emph{affordances}. Constructing such unified 3DSGs, however, is fundamentally challenging for two reasons. First, \textit{part-level open-vocabulary grounding is intrinsically harder than object-level grounding}: parts are smaller, more ambiguous, heavily viewpoint-dependent, and systematically under-represented in the predominant object-centric data distributions of modern foundation models~\cite{zhong2022regionclip, li2022grounded, lin2022learning, zhou2022detecting, liu2024grounding, ren2024grounded}---often producing object-level localization even with part-centric prompts (\cref{fig:intro}). Further, transferring robust part grounding into unconstrained \emph{3D} reconstruction pipelines remains brittle even with dedicated part datasets~\cite{he2022partimagenet,ramanathan2023paco}. Second, unification \emph{explodes the reasoning space}: once parts become nodes, naive relation prediction scales quadratically with the number of entities, quickly becoming computationally expensive and increasingly error-prone if one relies on unconstrained LLM inference over all candidate pairs---leading to spurious functional relations or missed dependencies.

To address these challenges, we propose \textbf{\textit{OP3DSG}}, the \textbf{O}pen-vocabulary \textbf{P}art-aware \textbf{3D} \textbf{S}cene \textbf{G}raph generation framework for real-world environments. OP3DSG makes unified graphs feasible by combining: (i) \emph{knowledge-guided control of the entity space} that expands and validates part candidates via object-part structure, (ii) \emph{part-aware multi-view 3D fusion} that preserves fine-grained geometry for small components across viewpoints, and (iii) \emph{geometry-anchored, verification-gated LLM reasoning} that scales relation/affordance inference without exhaustive node/edge enumeration while avoiding hallucinations. Our key insight is that \emph{geometry can act as a strong prior that makes open-vocabulary, part-aware reasoning tractable}. Instead of inferring a unified graph from scratch, we first build a geometry-initialized prior graph that captures stable spatial neighborhood structure and object-part hierarchy from multi-view fusion. We then use an LLM not as a free-form generator, but as a \emph{verification-gated refiner}: it proposes and validates only those relations and affordances that are plausible under geometric constraints, turning unstructured language reasoning into constrained refinement.

% Our approach combines (i) object-part knowledge guided open-vocabulary 2D detection module that expands object candidates into admissible part queries and suppresses spurious part proposals, (ii) a fine-grained part-level 3D fusion module that robustly lifts 2D part masks into 3D and preserves small and thin instances via geometric--photometric consistency, and (iii) a prior 3D scene graph construction stage that initializes spatial structure from geometry and enables scalable LLM-based reasoning under verification gates, avoiding hallucinations and expensive VLM-based node/edge enumeration.

To systematically evaluate the unified 3D scene graph generation task, we build the \textit{UniGraph3D} benchmark. UniGraph3D consolidates the function-centric 3DSG datasets~\cite{delitzas2024scenefun3d, zhang2025open} and augments them with annotations for interactive object-part structure, spatial and functional relations, and affordances. Our benchmark not only supports fair comparison across prior methods but also facilitates future work on fine-grained 3D scene understanding. Experiments show that our OP3DSG substantially improves part-level grounding and multi-level relation prediction in indoor environments, and that unified graphs serve as a strong perception backbone for downstream embodied tasks such as language-conditioned querying~\cite{szymanska2024space3d, gu2024conceptgraphs}, navigation~\cite{li2022remote}, and task planning~\cite{Agia2022TaskographyER, Rana2023SayPlanGL}.

In summary, we make three contributions:
\begin{itemize}[label=$\bullet$]
    \item \textbf{Problem Formulation.} We introduce the \textit{Unified 3D Scene Graph} generation task, which jointly models objects, interactive parts, spatial and functional relations, and affordances within a single representation, enabling fine-grained embodied reasoning. 
    \item \textbf{Model Design.} We propose \textit{OP3DSG}, the unified 3D scene graph generation framework integrating knowledge-guided part grounding, fine-grained 3D fusion, and geometry-anchored, verification-gated reasoning to mitigate hallucinations.
    \item \textbf{Benchmark Setup.} We introduce \textit{UniGraph3D}, a benchmark with consolidated datasets, enriched annotations, and evaluation protocols for systematic assessment of unified 3D scene graph generation.
\end{itemize}

\section{Related Work}

% Other works~\cite{chang2023context, werby2024hierarchical} further explore context-aware grounding and hierarchical open-vocabulary scene graph construction for downstream tasks.

\subsubsection{3D Scene Graph.}
3DSGs that encode semantic information can be categorized into two research directions: \textit{Spatial 3DSGs} and \textit{Functional 3DSGs}.

$(i)$ \textit{Spatial 3DSGs} represent objects as nodes and model spatial relationships between objects as edges. Early works~\cite{kim20193, wald2020learning, wu2021scenegraphfusion, armeni20193d, wu2023incremental, feng20233d} adopt this paradigm, constructing graphs based on geometric consistency and object detection from point clouds or RGB-D inputs. To support higher-level abstraction, hierarchical 3DSG methods~\cite{armeni20193d, Hughes2022HydraAR, Rosinol2021KimeraFS, chang2023hydra} organize scenes into multi-level graphs (\eg, building--room--object hierarchies), enabling scalable reasoning over large environments. With the emergence of 2D foundation models~\cite{radford2021learning, li2022grounded, liu2024grounding, ren2024grounded}, recent approaches~\cite{koch2024open3dsg, gu2024conceptgraphs, chang2023context, werby2024hierarchical} have extended spatial 3DSG generation to an open-vocabulary setting. Open3DSG~\cite{koch2024open3dsg} co-embeds geometric features with VLM embeddings, enabling zero-shot prediction of both object and relations. ConceptGraphs~\cite{gu2024conceptgraphs} proposes the training-free pipeline by leveraging 2D foundation models and multi-view association.  Despite this progress, existing methods remain largely object-centric and tend to struggle with small or thin structures, leading to incomplete recognition of fine-grained parts. As a result, spatial 3DSGs may overlook interactive parts that are critical in embodied scenarios~\cite{li2022remote, Agia2022TaskographyER, Rana2023SayPlanGL}.

$(ii)$ \textit{Functional 3DSGs} focus on modeling functional relations between objects or parts. As an emerging line of research, recent approaches~\cite{zhang2025open, rotondi2025fungraph} leverage VLMs and LLMs to encode functional knowledge by generating rich language descriptions for each node to capture interaction semantics. However, existing methods still exhibit limited part-level perception ability and do not encode spatial structure. Moreover, relying on VLM-based captioning to compensate for insufficient detection granularity introduces substantial computational overhead and prolongs the graph construction time.

Considering spatial and functional graphs in isolation yields incomplete scene representations. We therefore move toward Unified 3DSGs, providing a comprehensive representation that jointly models part-level nodes with spatial and functional relations.

\subsubsection{Part-level Perception.}
Beyond object-level recognition~\cite{li2022grounded, minderer2022simple, wu2023aligning}, part-level segmentation~\cite{Wang2015JointOA, Liu2020PartawarePN, li2022panoptic} aims to achieve fine-grained understanding by recognizing the constituent components of objects. With the emergence of large-scale benchmarks~\cite{he2022partimagenet, ramanathan2023paco, mo2019partnet, zhou2019semantic, meletis2020cityscapes} for common objects, research on part-level perception has expanded to more diverse object categories~\cite{wei2023ov, sun2023going, mo2019partnet}. Subsequent works have further extended part-level understanding into the 3D domain, either by lifting 2D predictions into 3D space~\cite{Liu2022PartSLIPLP, Zhou2023PartSLIPEL, Yang2024SAMPart3D, Ma2024FindAP, Geng2022GAPartNetCD} or by directly performing part segmentation on detailed point cloud representations~\cite{Yu2019PartNetAR, Wang2020FewShotLO, Wang2019TransformerF3}. Building upon recent advances in part-level perception and 2D–3D lifting techniques, we extend part-level understanding of common indoor objects to the 3D domain. Specifically, we project part-level predictions into 3D space and associate them with geometric structures to ensure spatial consistency. Based on these part-aware representations, we construct part-aware 3D scene graphs, enabling structured and fine-grained scene understanding.

\section{Problem Formulation}
Given a posed RGB-D sequence $\{I_t, D_t\}_{t=0}^{T}$, where $I_t$ denotes the RGB frame and $D_t$ the corresponding depth, our goal is to construct a unified 3D scene graph $\mathcal{G} = (\mathcal{V}, \mathcal{E})$ that encodes objects, parts, spatial relations, functional relations, and affordances. The vertex set $\mathcal{V} = \mathcal{V}^{o} \cup \mathcal{V}^{p}$ consists of two types of nodes: object nodes $\mathcal{V}^{o}$, representing physical objects in the environment, and interactive part nodes $\mathcal{V}^{p}$, representing sub-components of objects (\eg, \textit{button}, \textit{handle}, \textit{seat}). Each node $v \in \mathcal{V}$ has a set of affordance labels $\mathcal{A}^{v}$ (\eg, \textit{rotate}, \textit{pinch pull}, \textit{tip push}), which specifies feasible physical interactions. The edge set $\mathcal{E} = \mathcal{E}^{s} \cup \mathcal{E}^{f}$ encodes both spatial relations $\mathcal{E}^{s}$ (\eg, \textit{on}, \textit{next to}, \textit{part of}) and functional relations $\mathcal{E}^{f}$ (\eg, \textit{turn on/off}, \textit{control flow}, \textit{remote}). 

% $\mathcal{G} = (\mathcal{V}, \mathcal{E}, \mathcal{A})$
% Each node is further associated with an affordance label $\mathcal{A}$ (\eg, \textit{rotate}, \textit{pinch pull}), which specifies feasible physical interactions.

\section{Method}

\begin{figure*}[t]
    \centering
    \includegraphics[width=\linewidth]{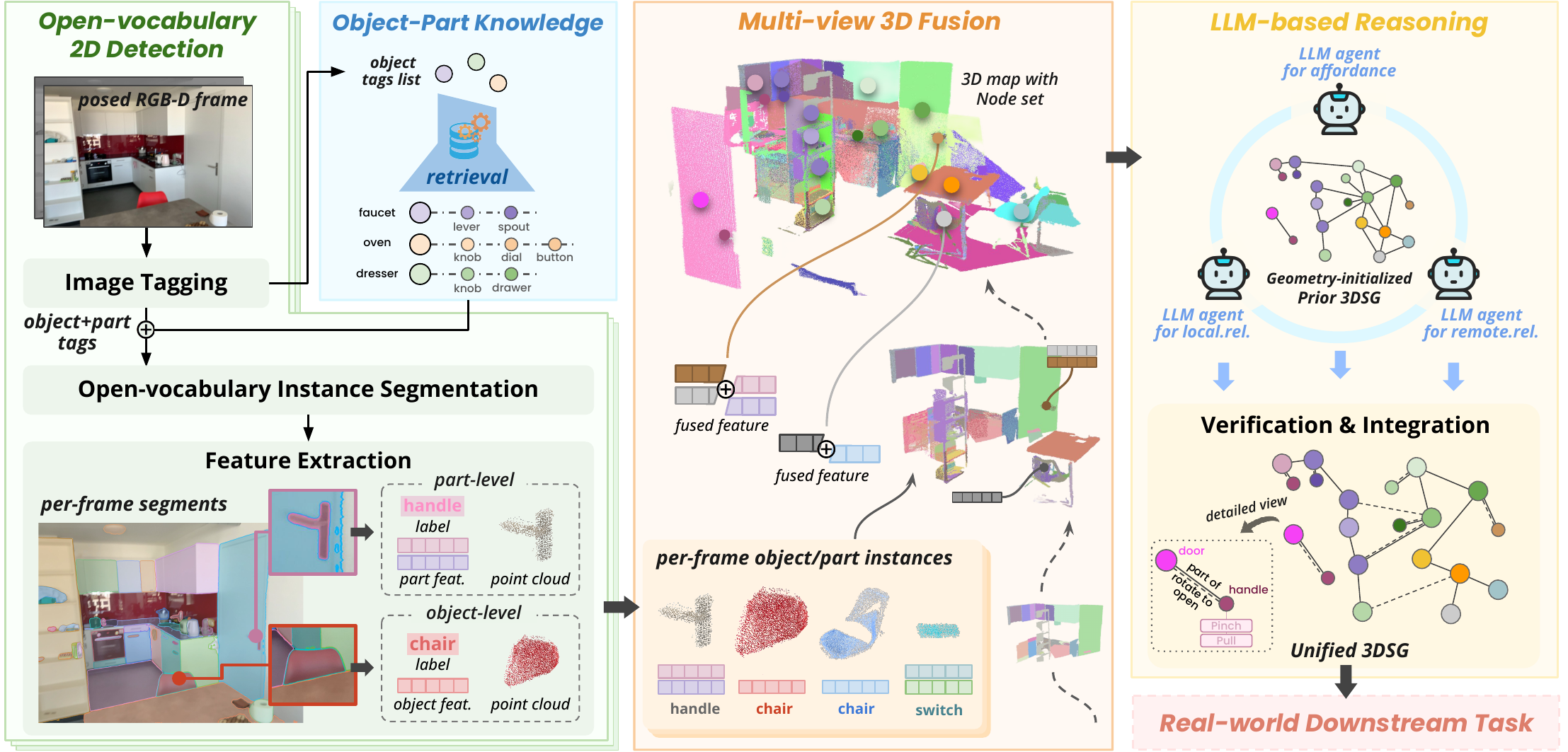}
    \caption{\textbf{Overview of the proposed framework.} Knowledge-guided 2D detection generates object and part instances that are incrementally fused into a global 3D map. The resulting nodes define a geometry-initialized prior graph, which is refined by LLM-based reasoning into the Unified 3DSG.}
    \label{fig:framework}
\end{figure*}

\subsection{Overview}
We propose OP3DSG, a model designed for part-level perception and unified 3D scene graph generation in an open-vocabulary manner. As illustrated in~\cref{fig:framework}, the framework consists of three main components: \textit{(i) object-part 2D detection}, \textit{(ii) multi-view 3D fusion}, and \textit{(iii) LLM-based reasoning}. 

In the object/part detection stage (\cref{sec:2D}), we leverage a foundation model pretrained on part-centric datasets to handle the challenge of fine-grained part recognition. By incorporating object–part knowledge, the model ensures comprehensive perception coverage and mitigates the omission of small or functionally important parts. The 3D fusion stage (\cref{sec:3D}) incrementally integrates the segmentation results from each frame into a global 3D map. To accurately merge small parts observed from multiple views, we propose a fine-grained fusion strategy that jointly considers geometric, chromatic, and semantic consistency. Finally, in the LLM-based reasoning stage (\cref{sec:LLM}), multiple LLM agents take the geometry-initialized prior 3DSG and each textual prompt as input. Unlike VLM-based methods, whose efficiency degrades as the number of objects grows, we adopt a language-only reasoning architecture with the geometry-anchored verification gate to avoid scalability bottlenecks while maintaining robust relational reasoning.

\subsection{Object-Part 2D Detection}
\label{sec:2D}
\subsubsection{Object/Part Candidates.}
Despite the success of foundation models in object detection, small objects or parts remain challenging. To address this gap, we introduce a simple strategy driven by object--part knowledge-guided control of the entity space. Given an RGB frame \(F_t\), an open-vocabulary image tagging model~\cite{zhang2024recognize} yields initial object candidates \(C^{o}_{t}\). 
A pre-built object--part knowledge base $\mathcal{K}$, empirically curated to cover common object–part relations in indoor scenes, is used to control the entity space. For each detected object label $c \in C^{o}_{t}$, we retrieve a predefined set of plausible interactive parts $\mathcal{K}(c)$ (\eg, \textit{oven} $\rightarrow$ \{\textit{oven button}, \textit{oven door}, \textit{oven handle}\}), forming the part candidate set $C^{p}_{t}=\bigcup_{c} \mathcal{K}(c)$. The final prompt list is constructed as $C_{t}=C^{o}_{t}\cup C^{p}_{t}$. This knowledge base guides part-level prompting without restricting the open-vocabulary nature of object recognition---for example, if \textit{door} tag is generated but its \textit{handle} or \textit{knob} is not recognized, we explicitly include the corresponding part tags before performing detection.

% A pre-built object--part knowledge, which expands the candidate set---for each detected object (\eg, \textit{oven}), it parses plausible parts (\eg, \textit{oven button},  \textit{oven door}, \textit{oven handle}) to form \(C^{p}_{t}\) and constructs the final candidate tag list \(C_{t}=C^{o}_{t}\cup C^{p}_{t}\). 
Furthermore, we adopt the open-vocabulary detection model~\cite{sun2023going} pre-trained on large-scale part-centric datasets~\cite{gupta2019lvis, ramanathan2023paco}. Prompted with \(C_t\), the detector outputs semantic label \(\{l_{t,i}\}_{i=1}^{N}\) and masks \(\{m_{t,i}\}_{i=1}^{N}\) for the \(N\) detected instances. Each mask is then back-projected to 3D and transformed into the global map, yielding the per-instance point cloud \(\text{p}_{t,n}\). Overall, the knowledge base provides a simple yet effective way to inject part-level knowledge to compensate for insufficient part recognition, while preserving the open-vocabulary nature of the detection modules.

% LVIS~\cite{gupta2019lvis} and PACO~\cite{ramanathan2023paco}

\subsubsection{Feature Extraction.}
For object association in 3D fusion, we extract region-level semantic features \(\{f^{s}_{t,i}\}_{i=1}^{N}\) from the recognition head of the detector, which align with the CLIP~\cite{radford2021learning} text embedding space. However, relying solely on semantic information is often inadequate for part-level association. Unlike objects that may comprise multiple heterogeneous materials, parts are typically composed of a consistent material. Motivated by this observation, we hypothesize that color and texture cues can provide more stable evidence for part association. Hence, we employ classical approaches~\cite{swain1991color, van2009learning, danelljan2015coloring, liang2015encoding} to compute color-distribution features \(f^{c}_{t,i}\) for part instances. For each part mask $m^{p}_{t,i}$, we apply white balancing to reduce illumination bias and define a valid pixel set $\Omega_{t,i}$ by excluding specular highlights. Specifically, we discard pixels that are simultaneously over-exposed and weakly chromatic, while keeping the remaining masked pixels:
% For each mask \(m^{p}_{t,i}\), white balancing is applied to reduce illumination bias and a valid pixel set \(\Omega_{t,i}\) is defined as:
% \begin{equation}
% \begin{aligned}
% \Omega_{t,i} = \{(x,y)\mid & \;m^p_{t,i}(x,y)=1, S(x,y)>\tau_S, \\
% & \neg(V(x,y)>\tau_V \land S(x,y)<0.1)\}\;,
% \end{aligned}
% \end{equation}
\begin{equation}
\begin{aligned}
\Omega_{t,i} = \{(x,y)\mid & \;m^p_{t,i}(x,y)=1, (S(x,y)>\tau_S) \land (V(x,y)<\tau_V)\}\;,
\end{aligned}
\end{equation}
% \begin{equation}
% \Omega_{t,i} = \{(x,y)\mid \;m^p_{t,i}(x,y)=1, S(x,y)>\tau_S, \neg(V(x,y)>\tau_V \land S(x,y)<0.1)\}\;,
% \end{equation}
where \(S\) and \(V\) denote the saturation and value components in the HSV color space, respectively. The saturation threshold \(\tau_S\) and the intensity threshold \(\tau_V\) jointly define a specular-like outlier region. The color-distribution feature \(f^{c}_{t,i}\) is concatenated with three normalized histogram vectors representing complementary color spaces: 
\begin{equation}
f^{c}_{t,i} = [\,H^{\mathrm{cn}}(\Omega_{t,i}),\, H^{\mathrm{ab}}(\Omega_{t,i}),\, H^{\mathrm{opp}}(\Omega_{t,i})\,]\;,
\end{equation}
where \(H^{\mathrm{cn}}\) is based on Color Names~\cite{van2009learning, danelljan2015coloring},  \(H^{\mathrm{opp}}\) on the Opponent color space~\cite{van2009evaluating, everts2013evaluation}, and \(H^{\mathrm{ab}}\) on the CIELAB \(a^{\ast}b^{\ast}\) space. 
% Additional formulas of each histogram can be found in the supplementary material.

\subsection{Multi-view 3D Fusion}
\label{sec:3D}
\subsubsection{Object Association.}
For every newly-detected object \({v}^{o}_{t,i} = (f^{s}_{t,i}, \text{p}_{t,i})\), we compute both geometric and semantic similarities against the existing objects in the map \(\{{v}^{o}_{t\!-\!1,j}\}_{j=1}^{J}\), where \({v}^{o}_{t\!-\!1,j} = (f^{s}_{t\!-\!1,j}, \text{p}_{t\!-\!1,j})\) might partially overlap in geometry. We define geometric similarity \(s^{g}\) as the nearest neighbor ratio (NNR), while semantic similarity \(s^{s}\) is the cosine similarity between L2-normalized descriptors and is affine-transformed to $[0,1]$ \cite{gu2024conceptgraphs}. The aggregated similarity \(S^{o}\) between the object \(i\) and \(j\) is defined as
\begin{equation}
s^{g}(i,j) = \textit{NNR}(\text{p}_{t,i}, \text{p}_{t,j}), \; s^{s}(i,j) = (f^{s\top}_{t,i} f^{s}_{t,j} + 1)/2,
% s^{g}(i,j) = \textit{NNR}(\text{p}_{t,i}, \text{p}_{t,j}), \; s^{s}(i,j) = \frac{f^{s\top}_{t,i} f^{s}_{t,j} + 1}{2},
\end{equation}
\begin{equation}
S^{o}(i,j) = s^{g}(i,j) + s^{s}(i,j)\;.
\end{equation}
We adopt a greedy online association strategy for incremental object fusion. For each newly detected object instance, we associate it with the existing objects that achieve the highest similarity. If the highest similarity is below a predefined threshold, a new object node is initialized.

\subsubsection{Part Association.} 
For every newly-detected part \({v}^{p}_{t,i} = (f^{s}_{t,i}, f^{c}_{t,i}, \text{p}_{t,i})\), 
we additionally compute color-distribution similarities against the existing parts in the map
\(\{{v}^{p}_{t\!-\!1,j}\}_{j=1}^{J}\), where \({v}^{p}_{t\!-\!1,j} = (f^{s}_{t\!-\!1,j}, f^{c}_{t\!-\!1,j}, \text{p}_{t\!-\!1,j})\) that partially overlap in geometry. The color-distribution similarity \(s^{c}\) between parts \(i\) and \(j\) is defined as a weighted sum of the chi-square distances \(d^{c}\) computed on each color histogram:
\begin{equation} % 
d^{c}(i,j) = w^c_{1} \cdot d^{\mathrm{cn}}(i,j)
          + w^c_{2} \cdot d^{ab}(i,j)
          + w^c_{3} \cdot d^{\mathrm{opp}}(i,j)\;,\;s^{c}(i,j)=\frac{1}{1 + d^{c}(i,j)},
\end{equation}
where \(d^{\mathrm{cn}}, d^{ab}, d^{\mathrm{opp}}\) denote the \(\chi^{2}\) distances between the corresponding histogram
\(H^{\mathrm{cn}}, H^{\mathrm{ab}}, H^{\mathrm{opp}}\), respectively, and \(w^{c}_{k}\) denotes the weight assigned to each component. The aggregated similarity \(S^{p}\) between part \(i\) and \(j\) is defined as
% \begin{equation}
% S^{p}(i,j) = w^p_{1} \cdot \big(s^{g}(i,j) + s^{s}(i,j)\big) + w^p_{2} \cdot s^{c}(i,j)\;,    
% \end{equation}
% where \(w^{p}_{n}\) denotes the weights for geometric/semantic and color similarity terms.
\begin{equation}
S^{p}(i,j) = w^p \cdot \big(s^{g}(i,j) + s^{s}(i,j)\big) + (1-w^p) \cdot s^{c}(i,j)\;,    
\end{equation}
where \(w^{p}\) denotes the weights for geometric/semantic and color-distribution similarity terms. We follow the same greedy assignment of object association to enable efficient streaming fusion without requiring a global optimal assignment.

\subsubsection{Feature Update.}
For each associated node \({v}^{o}_{t\!-\!1,j}\) and \({v}^{p}_{t\!-\!1,j}\), the corresponding features are updated incrementally with a new observation. We update the semantic feature by averaging the previous and current representations, while the geometric information by taking the union of point clouds by down-sampling to remove duplicate points \cite{gu2024conceptgraphs}: 
\begin{equation}
f^{s}_{t,j} \leftarrow \frac{n_{t\!-\!1,j} \cdot f^{s}_{t\!-\!1,j} + f^{s}_{t,i}}{n_{t\!-\!1,j} + 1}\;, \quad \text{p}_{t,j} \leftarrow \text{p}_{t\!-\!1,j} \,\cup\, \text{p}_{t,i}\;,
\end{equation}
where \(n_{t\!-\!1,j}\) denotes the number of detections previously associated with \({v}^{o}_{t\!-\!1,j}\) or \({v}^{p}_{t\!-\!1,j}\).
For the color-distribution feature, each histogram is updated according to the exponential moving average (EMA) rule. For each histogram \(H^{(k)}\) with \(k \in \{\mathrm{cn}, \mathrm{ab}, \mathrm{opp}\}\), the update is defined as
{\small
\begin{equation}
\tilde{H}^{(k)}_{t} = (1-\alpha)\,H^{(k)}_{t-1} + \alpha\,\widehat{H}^{(k)}_{t}, \quad
H^{(k)}_{t} = \frac{\tilde{H}^{(k)}_{t}}{\|\tilde{H}^{(k)}_{t}\|_{1} + \varepsilon}\;,    
\end{equation}}
where \(\alpha \in (0,1)\) is a smoothing factor, $\|\cdot\|_{1}$ denotes the L$^{1}$ norm, and $\varepsilon>0$ is a small constant for numerical stability.

\subsection{LLM Reasoning}
\label{sec:LLM}

\subsubsection{Prior 3D Scene Graph Generation.}
To compensate for the limited spatial perception capability of LLMs, we first construct a prior graph instead of directly encoding raw 3D coordinate information of nodes into the input prompt. From the final reconstructed 3D map \(\{v_{T,j}\}_{j=1}^{J}\), each node \(v_{T,j} = (l_{T,j}, \text{p}_{T,j})\) consists of a semantic label \(l_{T,j}\) and a point cloud \(\text{p}_{T,j}\). The label \(l_{T,j}\) is determined by a voting scheme that selects the most frequent detection label, while \(\text{p}_{T,j}\) is used to compute the 3D center coordinates and bounding box size of the node.

To construct edges, we treat nodes whose 3D centers are within 1 meter as neighbors, \ie, edge candidates. For inter-object pairs, we assign spatial relations (\eg, \textit{on}, \textit{next to}) based on their relative geometric configurations. To model intra-object connectivity, we further apply semantic proximity matching: for part nodes spatially-close to multiple object nodes, we retain only the edge whose connected object label is semantically consistent with the part label---\ie, when the object label text is contained within the part label text---and prune the remaining connections. Such a lexical heuristic is well-aligned with the knowledge-guided control, which generates part candidates in an object-conditioned form (\eg, \textit{cabinet handle})---promoting coherent object--part hierarchies while suppressing ambiguous attachments.

\subsubsection{Geometry-anchored LLM-based Reasoning.}

Given the prior 3DSG, we employ multiple LLM agents in parallel for reasoning. This multi-agent design decomposes the reasoning space---mitigating spurious predictions and explicitly separating short-range structural reasoning from long-range semantic reasoning. Each agent follows a step-wise prompt inspired by Chain-of-Thought~\cite{wei2022chain}. Before each reasoning, all agents \emph{independently} refine spatial relations in the prior graph: they validate inter-/intra-object relations by checking consistency with node labels, bounding-box extents, and 3D center coordinates derived from the prior 3DSG. This geometry-anchored verification filters spatially implausible relations and yields a physically consistent context for subsequent reasoning, without exhaustively enumerating all node pairs. The three agents are:

\textit{(i) Local relation reasoning agent} infers local functional relations, which typically arise between physically connected object--part pairs (\eg, \textit{knob -- pull to open -- cabinet}). Rather than iterating over all $O(N^2)$ node pairs, the agent conditions on the prior 3DSG and reasons over the spatially associated object--part pairs in a single prompt, ensuring scalability as the number of nodes increases.

\textit{(ii) Remote relation reasoning agent} handles functionally related but spatially distant entity pairs (\eg, \textit{TV -- turn on/off -- remote control}) by leveraging the LLM's pretrained world knowledge. To reduce the LLM's burden, the agent selects a set of remotable candidates from the prior graph and then predicts relation labels among these candidates, instead of considering all possible pairs.

\textit{(iii) Affordance reasoning agent} predicts feasible physical interactions for each node (\eg, \textit{handle -- pinch pull}, \textit{bottle -- contain}). For part nodes, affordance inference additionally conditions on the connected object label and geometric attributes, enabling context-aware and physically grounded predictions.

% After parallel reasoning, we apply majority voting: only relations consistently validated by at least two agents are retained. Further, the local/remote functional relations and affordances---disjoint components---are merged without conflict into a unified 3DSG. 

\section{Experiment}
\subsection{UniGraph3D}

To evaluate unified 3D scene graph generation, we propose \emph{UniGraph3D} by consolidating and extending the FunGraph3D \cite{zhang2025open} and SceneFun3D$^\dagger$ \cite{delitzas2024scenefun3d, zhang2025open} datasets under a unified annotation schema. Rather than merely merging existing resources, our human annotators carefully re-examine the annotations and identify substantial labeling inconsistencies and ambiguities (\eg, \textit{“rotate to adjust the setting”} vs. \textit{“rotate or press to adjust the setting”}). To address these issues, we consolidate ambiguous labels into a single canonical form, thereby enabling fair and consistent comparison across methods.

% Beyond annotation refinement, UniGraph3D significantly expands the original datasets by introducing explicit object–part spatial relations and comprehensive affordance annotations for each node. As a result, UniGraph3D supports systematic assessment of part-aware perception and multi-level relational reasoning, which are previously evaluated in isolation. As reported in \cref{tab:stat}, UniGraph3D provides a coherent and reliable testbed for unified 3D scene graph generation. Detailed statistics and procedures are in the supplementary material.

Beyond annotation refinement, UniGraph3D significantly expands the original datasets by introducing explicit object–part spatial relations and comprehensive affordance annotations for each node. Human annotators construct object–part connectivity based on the object and part labels with their coordinates. The affordances are assigned by aligning with existing work~\cite{delitzas2024scenefun3d} while considering the geometric characteristics of each node, resulting in nine categories. As a result, UniGraph3D supports systematic assessment of part-aware perception and multi-level relational reasoning, which were previously evaluated in isolation. As reported in \cref{tab:stat}, UniGraph3D provides a coherent and reliable testbed for unified 3D scene graph generation. Detailed statistics and procedures are provided in the supplementary material.

\begin{table}[t]
\centering
\tablesize
\captionsetup{width=1\linewidth}
\renewcommand{\arraystretch}{1.1}
\setlength{\aboverulesep}{1pt}
\setlength{\belowrulesep}{1pt}
\caption{\textbf{Comparison of 3DSG evaluation datasets.} We compare the number of objects, parts, spatial/functional relations, and affordances. The sign $\dagger$ denotes the post-processed dataset~\cite{delitzas2024scenefun3d} by additionally annotated functional relation~\cite{zhang2025open}.}
\setlength{\tabcolsep}{5.5pt}
\begin{tabular}{l c c c c c}
\toprule
\multicolumn{1}{c}{\textbf{Dataset}} &
\multicolumn{1}{c}{\hdr{Object Node}} &
\multicolumn{1}{c}{\hdr{Part Node}} &
\multicolumn{1}{c}{\hdr{Spat. Edge}} &
\multicolumn{1}{c}{\hdr{Func. Edge}} &
\multicolumn{1}{c}{\hdr{Affordance}}
\\
\midrule
ReplicaSSG~\cite{hou2025fross}  & 786 & 0 & 309 & 0 & 0 \\
FunGraph3D~\cite{zhang2025open} & 146 & 201 &  0 & 228 & 0 \\
SceneFun3D$^\dagger$~\cite{delitzas2024scenefun3d,zhang2025open}  & 105 &  212 & 0 & 195 & 212 \\
\hdashline[.4pt/2pt]
\rowcolor{ourgray}
\textbf{UniGraph3D (Ours)}      & 251 & 414 & 357 & 425 & 554  \\[-1pt] 
\bottomrule
\end{tabular}
\label{tab:stat}
\end{table}

\subsection{Settings}  
\subsubsection{Metrics.}
For node evaluation, we follow a retrieval-based protocol~\cite{zhang2025open, koch2024open3dsg}. A predicted node is considered correct if it has a non-zero 3D IoU with a ground-truth node and the corresponding ground-truth label ranks within the top-$K$ candidates based on cosine similarity between CLIP \cite{radford2021learning} embeddings of predicted and ground-truth labels. Node recall is computed as the proportion of ground-truth nodes retrieved under this criterion. 

For edge evaluation, a correct triplet (\texttt{<node-edge-node>}) is counted as retrieved within top-$K$ only if all its components are individually retrieved within top-$K$ under the same node retrieval criterion, and the relation is matched via cosine similarity of phrase embeddings. We utilize Sentence-BERT \cite{reimers2019sentence} because the previous method \cite{koch2024open3dsg, zhang2025open} based on BERT \cite{devlin2018bert} lacked the capability to properly evaluate relational phrases. Affordance evaluation is also performed using Sentence-BERT embeddings, where recall is computed solely based on affordance phrase similarity, independent of node label. We additionally report recall for the associated node, which considers only node pairs without a relation label.

% \subsubsection{Implementation Details.} 
% We employ RAM~\cite{zhang2024recognize} for image tagging and VLPart~\cite{sun2023going} that pre-trained on the LVIS~\cite{gupta2019lvis} and PACO~\cite{ramanathan2023paco} datasets for open-vocabulary detection. For each detected region, we further utilize SAM~\cite{kirillov2023segment} to generate instance masks. For LLM-based reasoning, we use GPT-5~\cite{singh2025openai} (\textsc{gpt-5-2025-08-07}). In our experiments, we set $\tau_S=0.1$ and $\tau_V=0.9$ for visual part feature extraction. During the 3D fusion stage, the weighting coefficients $w^c_{1}$, $w^c_{2}$, and $w^c_{3}$ are set to 0.2, 0.4, and 0.4, respectively, while $w^p$ is set to 0.4.

% % \subsubsection{Baselines.} 
% For the baselines, we compare with representative open-vocabulary 3D scene graph generation models~\cite{gu2024conceptgraphs, zhang2025open}. As neither baseline is originally designed to construct unified 3DSGs, we extend ConceptGraph~\cite{gu2024conceptgraphs} with additional part-level and functional relation inference, and augment OpenFunGraph~\cite{zhang2025open} with spatial relation reasoning. Affordance inference is incorporated into both baselines. For fair and up-to-date comparisons, we report results using both the originally adopted GPT-4~\cite{achiam2023gpt} and the more recent GPT-5~\cite{singh2025openai} for their LLM backbone.

\subsection{Quantitative Results}
We compare our OP3DSG with representative open-vocabulary 3DSG generation models~\cite{gu2024conceptgraphs, zhang2025open}. As neither baseline is originally designed to construct unified 3DSGs, we extend ConceptGraph~\cite{gu2024conceptgraphs} with additional part-level and functional relation inference via LLM prompts, and augment OpenFunGraph~\cite{zhang2025open} with spatial relation reasoning using the same prompting strategy. Affordance inference is incorporated into both baselines. For fair and up-to-date comparisons, we report results using both the originally adopted GPT-4~\cite{achiam2023gpt} (\textsc{gpt-4-0613}) and the more recent GPT-5~\cite{singh2025openai} (\textsc{gpt-5-2025-08-07}) model, which serves as the LLM backbone of OP3DSG. We also include FunGraph~\cite{rotondi2025fungraph} and KeySG~\cite{werby2026keysg} in the comparison, both of which rely only on inter-/intra-object relations without explicitly modeling functional relations.

%%%%%%%%%%%%%%%%%%%%%%%%%%%%%%%%%%%%%%%%%%%%%%%%%%%%%%%%%%%%%%%%%%%%%%%%%%%%%%%%%%%%%%%%%%%%%%%%%%%%
%%%%%%%%%%%%%%%%%%%%%%%%%%%%%%%%%%%%%%%%%%%%%%%%%%%%%%%%%%%%%%%%%%%%%%%%%%%%%%%%%%%%%%%%%%%%%%%%%%%%
\begin{table}[t]
\caption{\textbf{Node evaluation on UniGraph3D dataset.} The sign $\ast$ denotes that an additional VLM/LLM prompt is used to infer the part node; ++ indicates the use of the recent GPT-5~\cite{singh2025openai} model for fair comparison.}
% $\ddagger$ means usage of OP3DSG 3D fusion results as input for fair comparison.
\label{tab:main_node}
\centering
\tablesize
\setlength{\tabcolsep}{7.5pt}
\renewcommand{\arraystretch}{1.1}
\setlength{\aboverulesep}{1pt}
\setlength{\belowrulesep}{1pt}
\begin{tabular}{l c c c c c c c c}
\toprule
\multicolumn{1}{c}{\multirow{2}{*}[-1.0ex]{\textbf{Model}}} &
\multicolumn{2}{c}{\hdr{Object Node}} &
\multicolumn{2}{c}{\hdr{Part Node}} &
\multicolumn{2}{c}{\hdr{Overall Node}} &
\multicolumn{2}{c}{\hdr{Affordance}} \\
\cmidrule(lr){2-3}\cmidrule(lr){4-5}\cmidrule(lr){6-7}\cmidrule(lr){8-9}
& \multicolumn{1}{c}{R@3} & \multicolumn{1}{c}{R@5}
& \multicolumn{1}{c}{R@3} & \multicolumn{1}{c}{R@5}
& \multicolumn{1}{c}{R@3} & \multicolumn{1}{c}{R@5}
& \multicolumn{1}{c}{R@3} & \multicolumn{1}{c}{R@5} \\
\midrule
% Open3DSG$^\ast$~\cite{koch2024open3dsg}
% & . & . & . & . & . & . & . & . \\
% Open3DSG$^\ast$$^\ddagger$~\cite{koch2024open3dsg}
% & . & . & . & . & . & . & . & . \\
% \hdashline[.4pt/2pt]
ConceptGraph$^\ast$~\cite{gu2024conceptgraphs}
& 45.4 & 51.8 & 17.6 & 21.5 & 28.1 & 32.9 & 2.3 & 3.8 \\
ConceptGraph$^\ast$++~\cite{gu2024conceptgraphs}
& 53.4 & 61.4 & 16.7 & 19.6 & 30.5 & 35.3 & 4.9 & 5.2 \\
\hdashline[.4pt/2pt]
FunGraph~\cite{rotondi2025fungraph}
& 68.5 & 73.7 & 28.3 & 39.1 & 43.5 & 52.2 & 40.3 & 40.8 \\
FunGraph++~\cite{rotondi2025fungraph}
& 68.9 & 73.7 & 28.5 & 40.3 & 43.8 & 52.9 & 40.4 & 40.8 \\
\hdashline[.4pt/2pt]
OpenFunGraph~\cite{zhang2025open}
& 56.2 & 64.5 & 49.3 & 51.0 & 51.9 & 56.1 & 13.7 & 16.1 \\
OpenFunGraph++~\cite{zhang2025open}
& 62.2 & 69.3 & 51.7 & 52.4 & 55.6 & 58.8 & 32.1 & 33.9 \\
\hdashline[.4pt/2pt]
KeySG~\cite{werby2026keysg}
& 74.1 & 76.5 & 52.4 & 54.1 & 60.6 & 62.6 & 25.8 & 30.5 \\
\hdashline[.4pt/2pt]
\rowcolor{ourgray}
\textbf{OP3DSG (Ours)}
& \bfseries 84.9 & \bfseries 93.2 & \bfseries 83.6 & \bfseries 86.2 & \bfseries 84.1 & \bfseries 88.9 & \bfseries 76.7 & \bfseries 83.2 \\ [-1pt]
\bottomrule
\end{tabular}
\end{table}

%%%%%%%%%%%%%%%%%%%%%%%%%%%%%%%%%%%%%%%%%%%%%%%%%%%%%%%%%%%%%%%%%%%%%%%%%%%%%%%%%%%%%%%%%%%%%%%%%%%%
%%%%%%%%%%%%%%%%%%%%%%%%%%%%%%%%%%%%%%%%%%%%%%%%%%%%%%%%%%%%%%%%%%%%%%%%%%%%%%%%%%%%%%%%%%%%%%%%%%%%
\begin{table}[t]
\caption{\textbf{Edge evaluation on UniGraph3D dataset.} The sign $\ddagger$ denotes that an additional VLM/LLM prompt is used to infer the spatial relation. Assoc. Node refers to the node association metric.}
\label{tab:main_edge}
\centering
\tablesize
\setlength{\tabcolsep}{5.2pt}
\renewcommand{\arraystretch}{1.1}
\setlength{\aboverulesep}{1pt}
\setlength{\belowrulesep}{1pt}
\begin{tabular}{l c c c c c c c c c}
\toprule
% \multicolumn{1}{c}{\multirow{2}{*}[-1.0ex]{\textbf{Model}}} &
% \multicolumn{3}{c}{\scriptsize \textbf{Assoc. Node}} &
% \multicolumn{3}{c}{\scriptsize \textbf{Func. Triplet}} &
% \multicolumn{3}{c}{\scriptsize \textbf{Spat. Triplet}} \\
\multicolumn{1}{c}{\multirow{2}{*}[-1.0ex]{\textbf{Model}}} &
\multicolumn{3}{c}{\hdr{Assoc. Node}} &
\multicolumn{3}{c}{\hdr{Func. Triplet}} &
\multicolumn{3}{c}{\hdr{Spat. Triplet}} \\
\cmidrule(lr){2-4}\cmidrule(lr){5-7}\cmidrule(lr){8-10}
& \multicolumn{1}{c}{R@3} & \multicolumn{1}{c}{R@5} & \multicolumn{1}{c}{R@10}
& \multicolumn{1}{c}{R@3} & \multicolumn{1}{c}{R@5} & \multicolumn{1}{c}{R@10}
& \multicolumn{1}{c}{R@3} & \multicolumn{1}{c}{R@5} & \multicolumn{1}{c}{R@10} \\
% & \multicolumn{1}{c}{\scriptsize R@3} & \multicolumn{1}{c}{\scriptsize R@5} & \multicolumn{1}{c}{\scriptsize R@10}
% & \multicolumn{1}{c}{\scriptsize R@3} & \multicolumn{1}{c}{\scriptsize R@5} & \multicolumn{1}{c}{\scriptsize R@10}
% & \multicolumn{1}{c}{\scriptsize R@3} & \multicolumn{1}{c}{\scriptsize R@5} & \multicolumn{1}{c}{\scriptsize R@10} \\
\midrule
% Open3DSG$^\ast$~\cite{koch2024open3dsg}
% & . & . & . & . & . & . & . & . & . \\
% Open3DSG$^\ast$$^\ddagger$~\cite{koch2024open3dsg}
% & . & . & . & . & . & . & . & . & . \\
% \hdashline[.4pt/2pt]
ConceptGraph$^\ast$~\cite{gu2024conceptgraphs}
& 4.2 & 5.2 & 7.3 & 5.9 & 6.6 & 8.5 & 1.7 & 1.7 & 2.2 \\
ConceptGraph$^\ast$++~\cite{gu2024conceptgraphs}
& 7.7 & 8.7 & 14.5 & 11.5 & 12.7 & 19.8 & 2.8 & 3.1 & 7.3 \\
\hdashline[.4pt/2pt]
FunGraph~\cite{rotondi2025fungraph}
& 9.6 & 13.3 & 21.7 & 0.0 & 0.0 & 0.0 & 21.0 & 29.1 & 47.6 \\
FunGraph++~\cite{rotondi2025fungraph}
& 9.8 & 13.9 & 21.9 & 0.0 & 0.0 & 0.0 & 21.6 & 30.5 & 47.9 \\
\hdashline[.4pt/2pt]
OpenFunGraph$^\ddagger$~\cite{zhang2025open}
& 38.2 & 41.9 & 44.5 & 36.9 & 40.2 & 45.2 & 35.9 & 39.8 & 43.1 \\
OpenFunGraph$^\ddagger$++~\cite{zhang2025open}
& 45.1 & 48.7 & 50.9 & 41.9 & 45.9 & 51.8 & 44.5 & 48.2 & 49.9 \\
\hdashline[.4pt/2pt]
KeySG~\cite{werby2026keysg}
& 12.9 & 17.1 & 21.5 & 0.0 & 0.0 & 0.0 & 28.3 & 37.5 & 47.1 \\
\hdashline[.4pt/2pt]
\rowcolor{ourgray}
\textbf{OP3DSG (Ours)}
& \bfseries 53.1 & \bfseries 66.9 & \bfseries 76.1 & \bfseries 51.1 & \bfseries 64.5 & \bfseries 74.8 & \bfseries 52.4 & \bfseries 66.9 & \bfseries 77.3 \\[-1pt]
\bottomrule
\end{tabular}
\end{table}

\subsubsection{Node Evaluation.}
\Cref{tab:main_node} presents the results of the node recall on the unified 3DSG generation task. Our framework substantially enhances the overall node performance, with particular gains in part-level recognition, surpassing the strongest baseline by +31.2 points (R@3) and +32.1 points (R@5). This large margin is not merely a byproduct of stronger language models, but reveals a structural limitation of existing open-vocabulary baselines. While upgrading the VLM/LLM backbone moderately improves object-node recall, part-node performance either remains low---indicating that scaling foundation models alone cannot resolve the intrinsic difficulty of open-vocabulary part grounding. In particular, part-centric prompts often collapse into object-level localization, yielding confident object detections but weak or fragmented part representations. In contrast, OP3DSG explicitly tackles this bottleneck at the perception and fusion levels. As a result, parts are not treated as noisy in object detection, but as geometrically stable and semantically coherent entities in the reconstructed scene.

We further evaluate the affordance predictions for each node in \Cref{tab:main_node} and observe that OP3DSG significantly outperforms all baselines, achieving 76.7\% (R@3) and 83.2\% (R@5). Since affordance inference is conditioned primarily on the predicted object/part labels and relies on the LLM’s pretrained knowledge, the improved part-node recognition and stable object–part attachment contribute substantially to the performance gains. These results suggest that language-based affordance reasoning becomes substantially more effective without relying on per-node caption generation once correct interaction anchors---well-grounded parts and their associated objects---are established.

\subsubsection{Edge Evaluation.}
\Cref{tab:main_edge} presents the edge recall of our method and the baselines on the unified 3DSG generation task. Our framework substantially improves overall edge performance, with gains of 9.2\%p (R@3) in functional relations and 7.9\%p (R@3) in spatial relations compared to the strongest baseline. These improvements persist even when baselines are strengthened with additional prompting and newer LLMs, indicating that the advantage does not stem from stronger language priors alone, but from more effective structural constraints. In OP3DSG, the geometry-initialized prior graph restricts candidate reasoning to structurally plausible neighborhoods, while the verification-gated multi-agent LLM refinement enforces geometric consistency during relation inference. Consequently, the LLM operates as a constraint-aware refiner rather than an unconstrained edge generator, improving both precision by suppressing spurious edges and recall by reducing missed dependencies in a scalable manner.

%%%%%%%%%%%%%%%%%%%%%%%%%%%%%%%%%%%%%%%%%%%%%%%%%%%%%%%%%%%%%%%%%%%%%%%%%%%%%%%%%%%%%%%%%%%%%%%%%%%%
%%%%%%%%%%%%%%%%%%%%%%%%%%%%%%%%%%%%%%%%%%%%%%%%%%%%%%%%%%%%%%%%%%%%%%%%%%%%%%%%%%%%%%%%%%%%%%%%%%%%
\begin{table}[!t]
\caption{\textbf{Node Localization Sensitivity study on UniGraph3D dataset.}}
\label{tab:ablation1}
\centering
\tablesize
\setlength{\tabcolsep}{5.2pt}
\renewcommand{\arraystretch}{1.1}
\setlength{\aboverulesep}{1pt}
\setlength{\belowrulesep}{1pt}
\begin{tabular}{l c c c c c c}
\toprule
\multicolumn{1}{c}{\multirow{2}{*}[-0.9ex]{\textbf{Model}}} &
\multicolumn{2}{c}{\hdr{Object Node}} &
\multicolumn{2}{c}{\hdr{Part Node}} &
\multicolumn{2}{c}{\hdr{Overall Node}} \\
\cmidrule(lr){2-3}\cmidrule(lr){4-5}\cmidrule(lr){6-7}
& \multicolumn{1}{c}{$\text{IoU}_{\ge 0.10}$} & \multicolumn{1}{c}{$\text{IoP}_{\ge 0.25}$}
& \multicolumn{1}{c}{$\text{IoU}_{\ge 0.10}$} & \multicolumn{1}{c}{$\text{IoP}_{\ge 0.25}$}
& \multicolumn{1}{c}{$\text{IoU}_{\ge 0.10}$} & \multicolumn{1}{c}{$\text{IoP}_{\ge 0.25}$} \\
\midrule
% Open3DSG$^\ast$~\cite{koch2024open3dsg}
% & . & . & . & . & . & .  \\
% Open3DSG$^\ast$$^\ddagger$~\cite{koch2024open3dsg}
% & . & . & . & . & . & .  \\
% \hdashline[.4pt/2pt]
ConceptGraph$^\ast$~\cite{gu2024conceptgraphs}
& 30.7 & 27.5 & 4.1 & 1.2 & 14.1 & 11.1  \\
ConceptGraph++$^\ast$~\cite{gu2024conceptgraphs}
& 32.3 & 28.7 & 5.6 & 2.4 & 15.6 & 12.3  \\
\hdashline[.4pt/2pt]
FunGraph~\cite{rotondi2025fungraph}
& 49.8 & 57.8 & 15.7 & 8.9 & 28.6 & 27.4 \\
FunGraph++~\cite{rotondi2025fungraph}
& 49.0 & 57.8 & 14.5 & 9.7 & 27.5 & 27.8 \\
\hdashline[.4pt/2pt]
OpenFunGraph~\cite{zhang2025open}
& 38.2 & 30.3 & 15.0 & 5.8 & 23.8 & 15.0  \\
OpenFunGraph++~\cite{zhang2025open}
& 43.4 & 35.9 & 15.0 & 6.0 & 25.7 & 17.3  \\
\hdashline[.4pt/2pt]
KeySG~\cite{werby2026keysg}
& 50.6 & 59.4 & \bfseries 30.4 & \bfseries 23.4 & 38.0 & \bfseries 37.0 \\
\hdashline[.4pt/2pt]
\rowcolor{ourgray}
\textbf{OP3DSG (Ours)}
& \bfseries 63.7 & \bfseries 61.8 & 25.8 & 13.0 & \bfseries 40.2 & 31.4 \\[-1pt]
\bottomrule
\end{tabular}
\end{table}

\begin{table}[!t]
\caption{\textbf{Ablation study on UniGraph3D dataset.} }
\label{tab:ablation2}
\centering
\tablesize
\setlength{\tabcolsep}{4.5pt}
\renewcommand{\arraystretch}{1.1}
\setlength{\aboverulesep}{1pt}
\setlength{\belowrulesep}{1pt}
\begin{tabular}{l c c c c c c c c c c}
\toprule
\multicolumn{1}{c}{\multirow{2}{*}[-0.9ex]{\textbf{Model}}} &
\multicolumn{2}{c}{\hdr[0.88]{\scriptsize{Object Node}}} &
\multicolumn{2}{c}{\hdr[0.88]{\scriptsize{Part Node}}} &
\multicolumn{2}{c}{\hdr[0.88]{\scriptsize{Func. Triplet}}} &
\multicolumn{2}{c}{\hdr[0.88]{\scriptsize{Spat. Triplet}}} &
\multicolumn{2}{c}{\hdr[0.88]{\scriptsize{Affordance}}} \\
\cmidrule(lr){2-3}\cmidrule(lr){4-5}\cmidrule(lr){6-7}\cmidrule(lr){8-9}\cmidrule(lr){10-11}
& \multicolumn{1}{c}{R@3} & \multicolumn{1}{c}{R@5}
& \multicolumn{1}{c}{R@3} & \multicolumn{1}{c}{R@5}
& \multicolumn{1}{c}{R@3} & \multicolumn{1}{c}{R@5}
& \multicolumn{1}{c}{R@3} & \multicolumn{1}{c}{R@5}
& \multicolumn{1}{c}{R@3} & \multicolumn{1}{c}{R@5} \\
\midrule
\rowcolor{ourgray} \textbf{OP3DSG (Ours)}
& \bfseries 84.9 & \bfseries 93.2 & \bfseries 83.6 & \bfseries 86.2 & \bfseries 51.1 & \bfseries 64.5 & \bfseries 52.4 & \bfseries 66.9 & \bfseries 76.7 & \bfseries 83.2  \\
\; \scriptsize{w/o} \tablesize{knowledge}
& 76.9 & 86.5 & 45.9 & 48.3 & 15.5 & 21.2 & 12.6 & 18.8 & 30.7 & 33.6 \\
\; \scriptsize{w/o} \tablesize{color-dist.\;feat.}
& 84.9 & 93.2 & 79.0 & 81.6 & 40.2 & 52.0 & 41.7 & 49.9 & 64.8 & 70.8 \\
\; \scriptsize{w/o} \tablesize{prior graph}
& 84.9 & 93.2 & 83.6 & 86.2 & 47.1 & 58.1 & 47.6 & 59.1 & 74.7 & 80.9 \\
\; \scriptsize{w/o} \tablesize{split LLM}
& 84.9 & 93.2 & 83.6 & 86.2 & 20.7 & 27.8 & 19.3 & 27.7 & 40.8 & 45.0 \\
\bottomrule
\end{tabular}
\end{table}

\subsubsection{Node Localization Sensitivity.} To analyze geometric localization accuracy, we evaluate node retrieval under strict overlap constraints, as reported in \Cref{tab:ablation1}. Unlike our main retrieval protocol that uses a non-zero 3D IoU as a minimal spatial consistency filter, we set each threshold to 0.10 for IoU and 0.25 for IoP (Intersection over Prediction)~\cite{Chen2024INViTEIA, werby2026keysg} when computing R@3. IoP measures the fraction of the predicted 3D bounding box that lies within the ground-truth box, and primarily evaluates over-expansion or leakage of predictions beyond the true object extent. In contrast, IoU emphasizes overall overlap quality and is more sensitive to under-coverage and center misalignment, particularly when only partially observed surfaces are fused. OP3DSG demonstrates strong overall performance under both criteria, indicating that its advantage extends beyond semantic retrieval to stricter geometric grounding. Nevertheless, OP3DSG still shows limited geometric localization accuracy for part nodes, reflecting the intrinsic difficulty of localizing small components in surface-based fusion.

\subsection{Ablation Study}
 \Cref{tab:ablation2} ablates key components of OP3DSG. Removing the knowledge-guided control of the entity space causes a substantial drop in part retrieval, accompanied by severe degradation in functional/spatial triplets and affordances. This indicates that incomplete part grounding propagates directly to higher-order graph reasoning. Disabling the color-distribution feature results in only a modest decrease in part recall, yet triplet and affordance performance declines more noticeably, suggesting that even minor part-association instability can be amplified at the relational level. Replacing the geometry-initialized prior graph with a flat list of node attributes reduces triplet and affordance recall, demonstrating that structured spatial context facilitates more reliable LLM-based inference even with identical node sets. Assigning all roles to a single LLM without decomposition leads to a sharp decline in relation and affordance, due not only to increased hallucinations but also to context overload from the enlarged input scope. Overall, these ablations confirm that unified 3DSG quality depends not only on node retrieval accuracy, but also on controlled part candidate expansion, stable part association, and structured geometric context for relation and affordance reasoning.

\begin{wrapfigure}[10]{r}{6cm}
    \centering
    \vspace{-8.2mm}
    \includegraphics[width=1\linewidth]{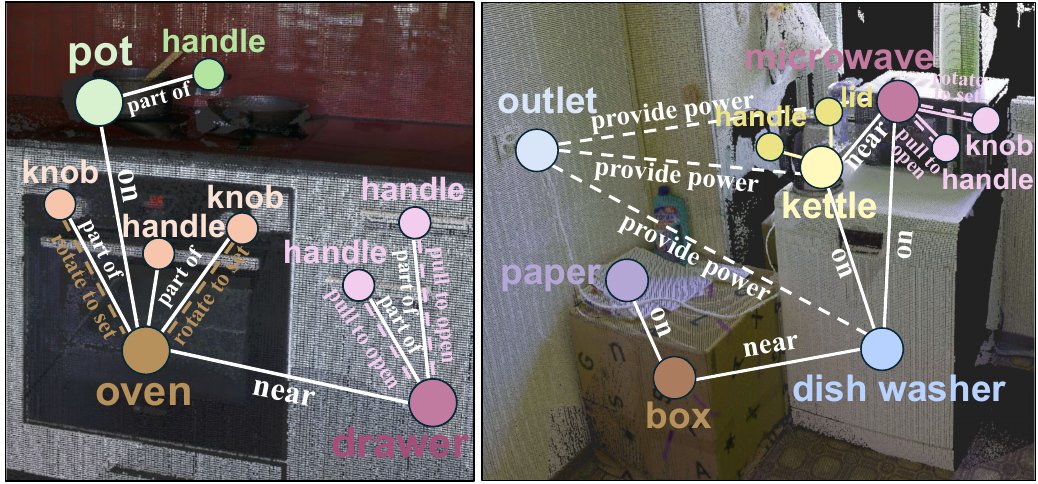}
    \caption{\textbf{Qualitative results.} We visualize the unified 3DSG in partial regions of two scenes.}
    \label{fig:qual}
\end{wrapfigure}
 \subsection{Qualitative Results} We visualize portions of the unified 3DSG in \Cref{fig:qual}. The graph jointly represents objects, their parts, and spatial or functional relations between them. For example, the oven and blender are connected to component nodes such as knobs, handles, and lids, while relational edges capture interactions like \textit{on}, \textit{near}, and \textit{provide power}, illustrating the fine-grained structure encoded in the unified 3DSG.

\section{Real-world Robot Applications}
\begin{figure}[t]
    \centering
    \includegraphics[width=1\linewidth]{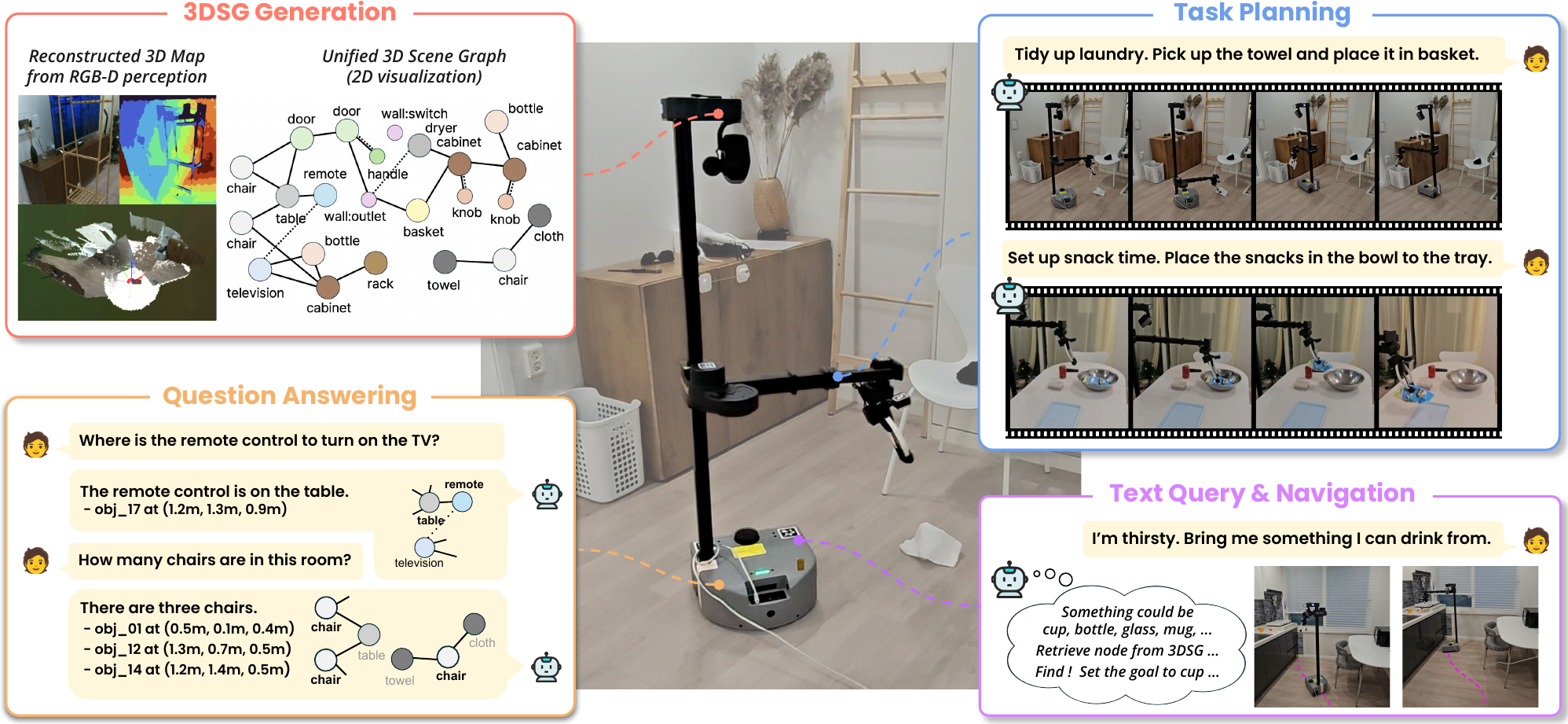}
    \caption{\textbf{Robot Applications using OP3DSG.} The generated 3DSGs can be used for various downstream tasks, including question answering, task planning, language-conditioned querying, and navigation. See the supplement for implementation details.}
    \label{fig:application}
\end{figure}

We demonstrate that OP3DSG directly supports diverse real-world embodied tasks (\cref{fig:application} and demo video). All experiments are conducted on the Stretch3 robot equipped with an Intel RealSense D435i RGB-D camera. During exploration, the robot incrementally reconstructs the environment and generates a unified 3DSG, which serves as a fine-grained representation for various downstream tasks. 
% This part-aware representation enables fine-grained grounding beyond object categories, allowing downstream modules to operate on small objects and functional components that are typically overlooked in object-only scene graphs. Detailed implementation of each module is described in the supplementary material.

% \subsubsection{Question Answering}

 % \textit{(i) \textbf{Question Answering}} is designed to assess the representational completeness and reasoning capacity of OP3DSG. In our pipeline, the LLM receives both the user’s query and the unified 3DSG as inputs, enabling language-guided reasoning directly over structured scene entities. By requiring counting, localization, relational inference, and part-level distinction, the QA task probes whether the unified 3DSG encodes sufficiently rich structural information to support interpretable spatial reasoning.
 \textit{(i) \textbf{Question Answering}} demonstrates OP3DSG supports language-guided reasoning over structured scene entities. Given a user query and the unified 3DSG, the LLM performs reasoning directly on the graph and handles questions involving counting, localization, relational inference, and part-level distinction---illustrating the fine-grained structural information in the unified 3DSG.

% \subsubsection{Task Planning}
 % \textit{(ii) \textbf{Task Planning}} is designed to evaluate whether OP3DSG can function as a perception backbone for grounding high-level language instructions into executable plans. Because the unified 3DSG maintains persistent object and part nodes with explicit spatial and relational edges, the planner can resolve references not only to whole objects but also to specific components (e.g., handles, containers, or functional buttons). This structured representation allows commands to be decomposed into part-aware subgoals and ordered according to spatial dependencies encoded in the graph. In contrast to object-only representations, OP3DSG supports more precise and controllable task specifications, where manipulation targets can be explicitly defined at the part level.
 \textit{(ii) \textbf{Task Planning}} demonstrates how OP3DSG grounds high-level language instructions into executable plans. The unified 3DSG maintains persistent object and part nodes with explicit spatial and relational edges, enabling the planner to resolve references to both objects and their components (\eg, handles, buttons). This representation allows instructions to be decomposed into part-aware subgoals and ordered according to spatial dependencies in the graph.

% \subsubsection{Text Query \& Navigation}
 % \textit{(iii) \textbf{Text Query}} \textbf{\&} \textit{\textbf{Navigation}} is designed to evaluate whether OP3DSG can serve as a language-conditioned spatial retrieval backbone under open-ended user intents. Unlike question answering, where the queried object category is explicitly specified, this task involves user requests that may not contain explicit object names but instead describe functional needs or situational intentions. Given a user query and the unified 3DSG, the system grounds such underspecified or open-vocabulary requests by retrieving candidate nodes at both object and part levels. Once a target node is selected, its 3D location is used to generate a navigation goal, allowing the robot to navigate toward the intended entity.
\textit{(iii) \textbf{Text Query}} \textbf{\&} \textit{\textbf{Navigation}} demonstrates language-conditioned spatial retrieval with OP3DSG under open-ended user intents. Unlike question answering, queries may not specify explicit object names but instead express functional needs or situational intentions. The system retrieves candidate nodes from the unified 3DSG at both object and part levels and uses the selected node’s 3D location to generate a navigation goal.

\section{Conclusion}
We introduced Unified 3D Scene Graph generation, aiming to represent indoor environments with a single compositional graph that jointly captures objects, parts, spatial/functional relations, and affordances. To make this challenging open-vocabulary setting tractable, we proposed OP3DSG, which couples knowledge-guided control of the entity space, part-aware multi-view 3D fusion, and geometry-anchored, verification-gated LLM refinement to constrain relation and affordance inference. We also presented UniGraph3D, enabling systematic evaluation of part-aware perception and multi-level relational reasoning. The proposed framework achieves substantial gains over prior open-vocabulary baselines. We further demonstrate that OP3DSG can serve as a robust perception backbone for real-world robotic applications, including complex text querying, navigation, and task planning. We expect our formulation, benchmark, and method will facilitate future research on scalable, part-aware 3D scene understanding and its deployment in real-world embodied systems.

\section*{Acknowledgements}
This research was partly supported by the Institute of Information \& Communications Technology Planning \& Evaluation (IITP) grant funded by the Korea government (MSIT) (No. RS-2022-II220907); by the National Research Foundation of Korea (NRF) grant funded by the Korea government (MSIT) (No. NRF-2022R1C1C1009989); and by the National Research Council of Science \& Technology (NST) grant funded by the Korea government (MSIT) (No. GTL25041-000).

\bibliographystyle{splncs04}
\bibliography{main}

\appendix

\title{Supplementary Material for OP3DSG: Open-Vocabulary Part-Aware 3D Scene Graph Generation for Real-World Environments} 

\titlerunning{Supplementary Material for OP3DSG}

% % TODO FINAL: Replace with your author list. 
% % Include the authors' OCRID for the camera-ready version, if at all possible.
% \author{
% Yirum Kim\inst{1}\orcidlink{0009-0000-4494-1839} \and
% % Ue-Hwan Kim\inst{1}\orcidlink{0000-0003-2201-2988}
% Ue-Hwan Kim\inst{1}\thanks{\scriptsize Corresponding author.}\orcidlink{0000-0003-2201-2988}
% }
\author{}

\authorrunning{Y. Kim and UH. Kim}

\institute{}
% % TODO FINAL: Replace with your institution list.
% \institute{Gwangju Institute of Science and Technology (GIST), Gwangju, Republic of Korea\\
% \email{kimyirum@gm.gist.ac.kr, uehwan@gist.ac.kr}
% }

\maketitle

\section*{Table of Contents}
\vspace{-16pt}
\cmirule
\vspace{4pt}
\tocitem{\ref{appendix:AddResults}}{\textbf{Additional Experiments}}{\pageref{appendix:AddResults}}
\tocsubitem{\ref{appendix:effAnal}}{Efficiency Analysis}{\pageref{appendix:effAnal}}
\tocsubitem{\ref{appendix:replica}}{Generalization to Object-Centric 3DSGs}{\pageref{appendix:replica}}
\tocsubitem{\ref{appendix:perData}}{Per-Dataset Evaluation}{\pageref{appendix:perData}}
\tocsubitem{\ref{appendix:qual}}{Qualitative Results}{\pageref{appendix:qual}}
\tocitem{\ref{appendix:Implement}}{\textbf{Implementation Details}}{\pageref{appendix:Implement}}
\tocsubitem{\ref{appendix:impleModel}}{Framework}{\pageref{appendix:impleModel}}
% \tocsubitem{\ref{appendix:implePrompt}}{LLM Prompts}{\pageref{appendix:implePrompt}}
\tocsubitem{\ref{appendix:app}}{Real-World Applications}{\pageref{appendix:app}}
\tocitem{\ref{appendix:Dataset}}{\textbf{Dataset Details}}{\pageref{appendix:Dataset}}
\tocsubitem{\ref{appendix:refine}}{Relation Label Refinement}{\pageref{appendix:refine}}
\tocsubitem{\ref{appendix:annot}}{Affordance Annotation}{\pageref{appendix:annot}}
\cmirule
\vspace{-5pt}
% \vspace{8pt}

\section{Additional Experiments}
\phantomsection
\label{appendix:AddResults}

\setcounter{figure}{5}
\setcounter{table}{5}

\begin{table}[t]
\caption{\textbf{Efficiency Comparison.} The peak GPU memory is measured during the 2D detection stage, and the presence of VLM/LLM calls required for graph construction is indicated. $N$ denotes the number of detected nodes.}
\label{sup:eff}
\centering
\scriptsize
\setlength{\tabcolsep}{15pt}
\renewcommand{\arraystretch}{1.3}
\setlength{\aboverulesep}{1pt}
\setlength{\belowrulesep}{1pt}
\begin{tabular}{lccc}
\toprule
\textbf{\qquad Model} & \textbf{Peak GPU Mem.} & \textbf{VLM Calls} & \textbf{LLM Calls} \\
\midrule
ConceptGraph~\cite{gu2024conceptgraphs} & 11.43 GB & $10N$ & $> N$ \\
OpenFunGraph~\cite{zhang2025open} & 11.90 GB & $\gg 10N$ & $> N$ \\
\hdashline[.4pt/2pt]
\rowcolor{ourgray}
\textbf{OP3DSG (Ours)}
& \bfseries 8.08 GB & \bfseries - & \bfseries $> 3$ \\[-1pt]
\bottomrule
\end{tabular}
\end{table}

\subsection{Efficiency Analysis}
\label{appendix:effAnal}

To provide a comprehensive analysis of efficiency, \Cref{sup:eff} reports the peak GPU memory usage during the 2D perception stage and the number of VLM/LLM calls required for 3D scene graph construction. Since all compared methods are training-free, the reported memory corresponds to the peak GPU usage during inference. The baselines rely on VLM-based caption generation to incorporate visual evidence for each detected node. Specifically, ConceptGraph~\cite{gu2024conceptgraphs} selects the top 10 views in which each node is observed and invokes VLM to generate captions for each view. The resulting captions are then refined using LLM on a per-node basis, after which the refined captions are used as input to an additional LLM call to construct the final scene graph. As a result, the number of VLM calls is approximately $10N$, where $N$ denotes the number of nodes in the scene, while the number of LLM calls is greater than $N$. OpenFunGraph~\cite{zhang2025open} follows a similar procedure for node caption generation, invoking the VLM for the top 10 views of each node and subsequently refining the captions using LLM. In addition, to infer object--part relations, OpenFunGraph generates extra VLM captions for images containing object--part candidates. Consequently, the total number of VLM calls becomes substantially larger than $10N$.

In contrast, the proposed method improves the perception stage to enable reliable part-level recognition without requiring VLM-based caption generation. As a result, graph construction is performed using only a small number of LLM-based reasoning steps. The baseline methods, however, additionally employ LLaVA~\cite{liu2023visual} as VLM, whose GPU memory consumption is not included in the reported peak GPU memory. Therefore, the reported memory values for the baselines should be regarded as conservative, and their actual end-to-end inference memory usage can be higher.

%   All signs follow the same notation as in \Cref{sup:node} and \Cref{sup:edge}. Assoc. Node refers to the node association metric. 
\begin{table}[t]
\caption{\textbf{Node/Edge evaluation on ReplicaSSG~\cite{hou2025fross} dataset.} }
\label{sup:rep}
\centering
\tablesize
\setlength{\tabcolsep}{11.3pt}
\renewcommand{\arraystretch}{1.1}
\setlength{\aboverulesep}{1pt}
\setlength{\belowrulesep}{1pt}
\begin{tabular}{l c c c c c c}
\toprule
\multicolumn{1}{c}{\multirow{2}{*}[-1.0ex]{\textbf{Model}}} &
\multicolumn{2}{c}{\hdr{Object Node}} &
\multicolumn{2}{c}{\hdr{Assoc. Node}} &
\multicolumn{2}{c}{\hdr{Spat. Triplet}} \\
\cmidrule(lr){2-3}\cmidrule(lr){4-5}\cmidrule(lr){6-7}
& \multicolumn{1}{c}{R@3} & \multicolumn{1}{c}{R@5}
& \multicolumn{1}{c}{R@3} & \multicolumn{1}{c}{R@5}
& \multicolumn{1}{c}{R@3} & \multicolumn{1}{c}{R@5} \\
\midrule
ConceptGraph$^\ast$~\cite{gu2024conceptgraphs}
& 48.7 & 51.8 & 18.4 & 21.4 & 9.7 & 11.7 \\
ConceptGraph$^\ast$++~\cite{gu2024conceptgraphs}
& 50.8 & 54.3 & 25.9 & 28.8 & 12.0 & 13.9 \\
\hdashline[.4pt/2pt]
OpenFunGraph$^\ddagger$~\cite{zhang2025open}
& 52.2 &  55.9 &  30.7 &  32.4 &  13.3 &  15.2  \\
OpenFunGraph$^\ddagger$++~\cite{zhang2025open}
& 53.6 &  56.9 &  39.8 &  41.4 &  22.7 &  23.0  \\
\hdashline[.4pt/2pt]
\rowcolor{ourgray}
\textbf{OP3DSG (Ours)}
& \bfseries 59.7 & \bfseries 63.6 & \bfseries 56.6 & \bfseries 63.4 & \bfseries 30.7 & \bfseries 31.1 \\ [-1pt]
\bottomrule
\end{tabular}
\end{table}

\subsection{Generalization to Object-Centric 3DSGs}
\label{appendix:replica}
To demonstrate that the proposed model can generalize to conventional object-centric 3DSG generation, we additionally evaluate our method on the ReplicaSSG dataset, as reported in \Cref{sup:rep}. ReplicaSSG contains 11 indoor scenes annotated with 9 types of spatial relations. Even under this object-centric setting, OP3DSG consistently outperforms the existing baselines across object node, associated node, and spatial triplet prediction.

A notable observation is that the performance gain is not confined to relation-level metrics, but already appears at the object node level. Although ConceptGraph is originally designed for object-level spatial 3DSG generation, it still shows relatively limited performance. This suggests that relation prediction is strongly affected by the quality of object perception itself---when object nodes are incomplete or inaccurate, downstream relation prediction also deteriorates. In contrast, the stronger object perception of OP3DSG leads to more reliable object grounding, which in turn improves associated node matching and spatial triplet prediction. These results indicate that the advantage of OP3DSG is not specific to the unified benchmark, but extends to conventional object-centric 3DSG generation as well.

% NODE EVAL 
\begin{table}[t]
\caption{\textbf{Node evaluation.} The sign $\ast$ denotes that an additional VLM/LLM prompt is used to infer the part node; ++ indicates the use of the recent GPT-5~\cite{singh2025openai} model for fair comparison.}
\label{sup:node}
\centering
\scriptsize
\setlength{\tabcolsep}{2.7pt}
\renewcommand{\arraystretch}{1.1}
\setlength{\aboverulesep}{1pt}
\setlength{\belowrulesep}{1pt}
\begin{tabular}{l c c c c c c c c c c c c}
\toprule
\multicolumn{1}{c}{\multirow{3}{*}[-1.5ex]{\textbf{Model}}} &
\multicolumn{6}{c}{\textbf{SceneFun3D$^\dagger$~\cite{delitzas2024scenefun3d, zhang2025open}}} &
\multicolumn{6}{c}{\textbf{FunGraph3D~\cite{zhang2025open}}} \\
\cmidrule(lr){2-7}\cmidrule(lr){8-13}
& \multicolumn{2}{c}{\hdr{Object}}
& \multicolumn{2}{c}{\hdr{Part}}
& \multicolumn{2}{c}{\hdr{Overall}}
& \multicolumn{2}{c}{\hdr{Object}}
& \multicolumn{2}{c}{\hdr{Part}}
& \multicolumn{2}{c}{\hdr{Overall}} \\
\cmidrule(lr){2-3}\cmidrule(lr){4-5}\cmidrule(lr){6-7}
\cmidrule(lr){8-9}\cmidrule(lr){10-11}\cmidrule(lr){12-13}
& R@3 & R@5
& R@3 & R@5
& R@3 & R@5
& R@3 & R@5
& R@3 & R@5
& R@3 & R@5 \\
\midrule
ConceptGraph$^\ast$~\cite{gu2024conceptgraphs} & 39.0 & 42.9 & 20.3 & 25.0 & 26.5 & 30.9 & 50.0 & 58.2 & 14.9 & 17.8 & 29.6 & 34.8 \\
ConceptGraph$^\ast$++~\cite{gu2024conceptgraphs} & 46.7 & 51.4 & 17.9 & 22.6 & 27.4 & 32.2 & 58.2 & 68.5 & 15.3 & 16.3 & 33.3 & 38.2 \\
\hdashline[.4pt/2pt]
OpenFunGraph~\cite{zhang2025open} & 65.7 & 69.5 & 63.7 & 64.6 & 64.4 & 66.2 & 49.3 & 61.0 & 34.2 & 36.6 & 40.5 & 46.8 \\
OpenFunGraph++~\cite{zhang2025open} & 66.7 & 68.6 & 66.0 & 67.0 & 66.2 & 67.5 & 58.9 & 69.9 & 36.6 & 37.1 & 46.0 & 50.9 \\
\hdashline[.4pt/2pt]
\rowcolor{ourgray}
\textbf{OP3DSG (Ours)} & \bfseries 89.5 & \bfseries 92.4 & \bfseries 87.7 & \bfseries 89.2 & \bfseries 88.3 & \bfseries 90.2 & \bfseries 81.5 & \bfseries 93.8 & \bfseries 79.2 & \bfseries 83.2 & \bfseries 80.2 & \bfseries 87.6 \\[-1pt]
\bottomrule
\end{tabular}
\end{table}

% EDGE EVAL
\begin{table}[t]
\caption{\textbf{Edge evaluation.} The sign $\ddagger$ denotes that an additional VLM/LLM prompt is used to infer the spatial relation. Assoc. refers to the node association metric.}
\label{sup:edge}
\centering
\scriptsize
\setlength{\tabcolsep}{2.5pt}
\renewcommand{\arraystretch}{1.1}
\setlength{\aboverulesep}{1pt}
\setlength{\belowrulesep}{1pt}
\begin{tabular}{l c c c c c c c c c c c c}
\toprule
\multicolumn{1}{c}{\multirow{3}{*}[-1.5ex]{\textbf{Model}}} &
\multicolumn{6}{c}{\textbf{SceneFun3D$^\dagger$~\cite{delitzas2024scenefun3d, zhang2025open}}} &
\multicolumn{6}{c}{\textbf{FunGraph3D~\cite{zhang2025open}}} \\
\cmidrule(lr){2-7}\cmidrule(lr){8-13}
& \multicolumn{2}{c}{\hdr{Assoc.}}
& \multicolumn{2}{c}{\hdr{Func.}}
& \multicolumn{2}{c}{\hdr{Spat.}}
& \multicolumn{2}{c}{\hdr{Assoc.}}
& \multicolumn{2}{c}{\hdr{Func.}}
& \multicolumn{2}{c}{\hdr{Spat.}} \\
\cmidrule(lr){2-3}\cmidrule(lr){4-5}\cmidrule(lr){6-7}
\cmidrule(lr){8-9}\cmidrule(lr){10-11}\cmidrule(lr){12-13}
& R@3 & R@5
& R@3 & R@5
& R@3 & R@5
& R@3 & R@5
& R@3 & R@5
& R@3 & R@5 \\
\midrule
ConceptGraph$^\ast$~\cite{gu2024conceptgraphs} & 4.5 & 5.1 & 6.2 & 6.7 & 1.2 & 1.2 & 4.0 & 5.4 & 5.7 & 6.5 & 2.0 & 2.0 \\
ConceptGraph$^\ast$++~\cite{gu2024conceptgraphs} & 6.7 & 7.6 & 10.3 & 10.8 & 2.5 & 2.5 & 8.5 & 9.6 & 12.6 & 14.3 & 3.1 & 3.6 \\ 
\hdashline[.4pt/2pt]
OpenFunGraph$^\ddagger$~\cite{zhang2025open} & 59.6 & 61.5 & 57.4 & 61.5 & 56.5 & 58.4 & 20.4 & 25.6 & 19.6 & 22.2 & 18.9 & 24.5 \\
OpenFunGraph$^\ddagger$++~\cite{zhang2025open} & 63.5 & 64.6 & 58.5 & 65.6 & 62.1 & 63.4 & 29.8 & 35.4 & 27.8 & 29.1 & 30.1 & 35.7 \\ 
\hdashline[.4pt/2pt]
\rowcolor{ourgray}
\textbf{OP3DSG (Ours)}
& \bfseries 64.0 & \bfseries 68.3 & \bfseries 62.1 & \bfseries 68.2 & \bfseries 61.5 & \bfseries 67.1 & \bfseries 43.9 & \bfseries 65.7 & \bfseries 41.7 & \bfseries 61.3 & \bfseries 44.9 & \bfseries 66.8 \\[-1pt]
\bottomrule
\end{tabular}
\end{table}

\subsection{Per-Dataset Evaluation}
\label{appendix:perData}
Since the proposed UniGraph3D is constructed by merging the SceneFun3D~\cite{delitzas2024scenefun3d} and FunGraph3D~\cite{zhang2025open} datasets under a unified annotation schema, the main paper reports results on the merged benchmark to provide a comprehensive comparison across methods. For completeness, we additionally report evaluation results on each dataset individually in \Cref{sup:node} and \Cref{sup:edge}. Since the two source datasets differ in label composition, relation distribution, and scene characteristics, this per-dataset evaluation helps verify that the observed improvements are not merely artifacts of the merged benchmark construction. All methods are evaluated using the same experimental settings as in the main paper, including identical detector configurations and LLM inference settings. %, to ensure a fair comparison.

Across both datasets, OP3DSG consistently outperforms the baselines in both node and edge prediction metrics. In particular, the improvement is most notable in part-level node recognition and functional relation prediction, which matches the trend observed on the merged UniGraph3D benchmark. This behavior is consistent with the design of the proposed framework---fine-grained node recognition benefits from the part-aware perception pipeline, while relation prediction is supported by geometry-anchored reasoning over the prior graph. The broadly preserved ranking across methods further suggests that the observed gains are not artifacts of dataset merging or annotation harmonization, but arise from the proposed model design. Overall, the consistent results on both datasets support the robustness and general applicability of our approach.

\clearpage
\begin{figure}[!ht]
\centering
\begin{subfigure}[t]{\linewidth}
    \centering
    \includegraphics[width=\linewidth]{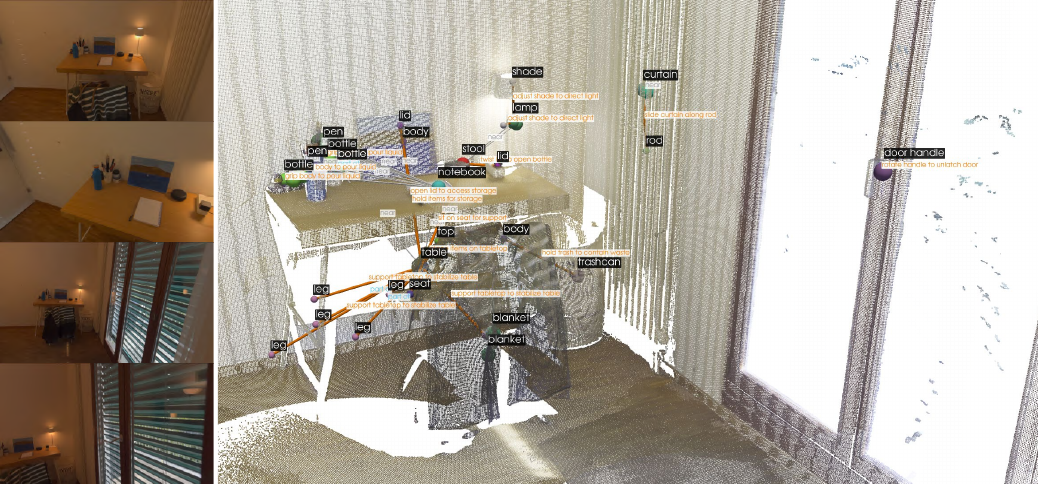}%{figures/qual_11bedroom_1.pdf}
    \caption{Unified 3DSG of the bedroom with a desk, chair, and lamp}
    \label{sup:qual_1}
\end{subfigure}
\begin{subfigure}[t]{\linewidth}
    \centering
    \includegraphics[width=\linewidth]{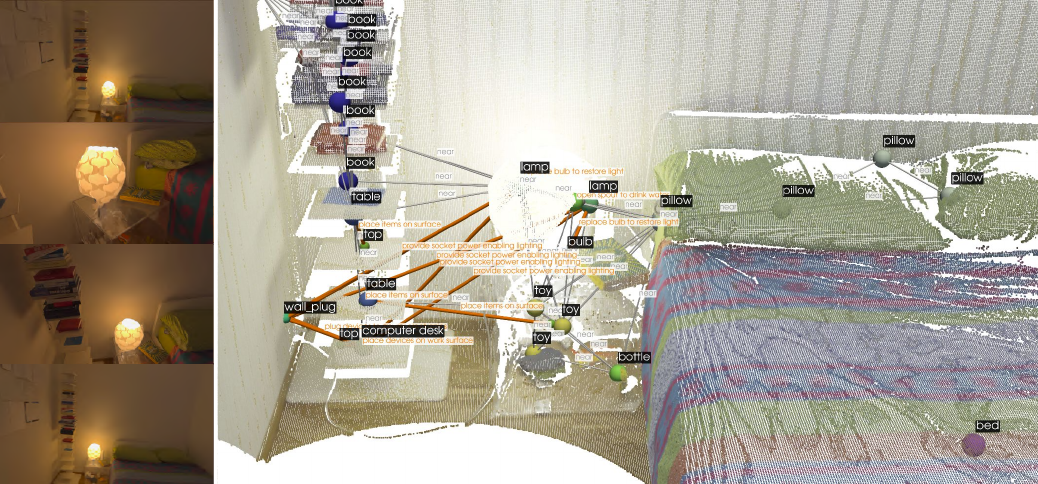}%{figures/qual_11bedroom_2.pdf}
    \caption{Unified 3DSG of the bedroom with a bed, a lamp, and books}
    \label{sup:qual_2}
\end{subfigure}
\begin{subfigure}[t]{\linewidth}
    \centering
    \includegraphics[width=\linewidth]{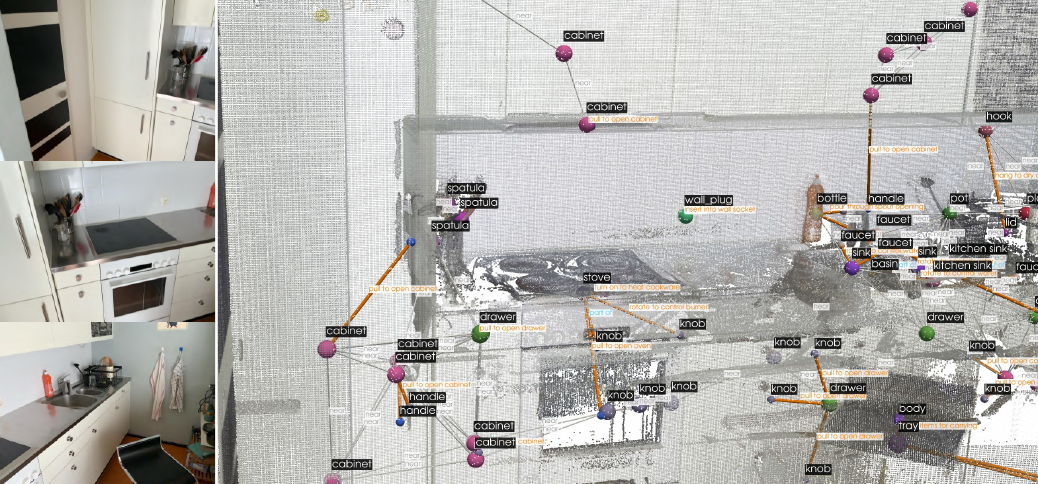}%{figures/qual_10kitchen_1.pdf}
    \caption{Unified 3DSG of the kitchen with a stove and cabinets}
    \label{sup:qual_3}
\end{subfigure}
% \caption{\textbf{Qualitative results of OP3DSG.}}
% \label{sup:qual}
\end{figure}
\begin{figure}[!ht]\ContinuedFloat
\centering
\begin{subfigure}[t]{\linewidth}
    \centering
    \includegraphics[width=\linewidth]{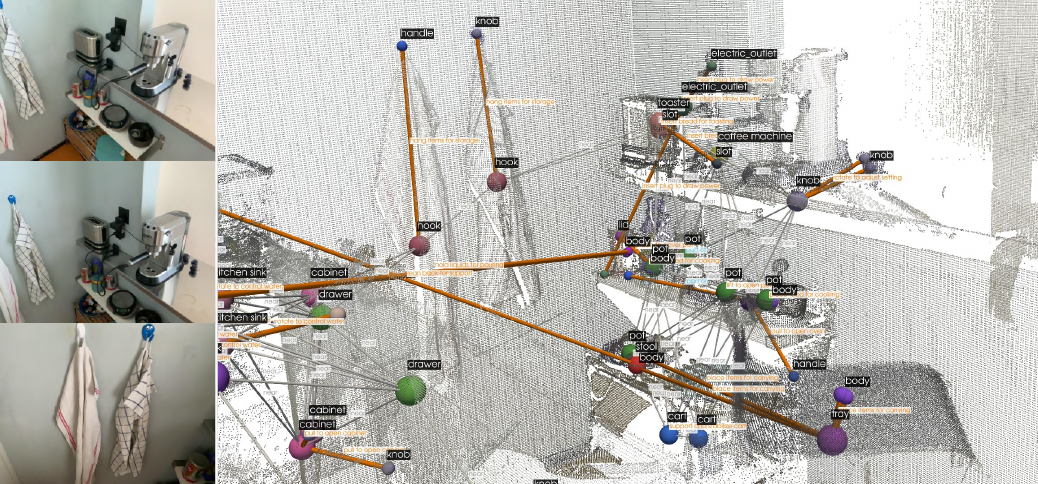}%{figures/qual_10kitchen_2.pdf}
    \caption{Unified 3DSG of the kitchen with some home appliances}
    \label{sup:qual_4}
\end{subfigure}
\begin{subfigure}[t]{\linewidth}
    \centering
    \includegraphics[width=\linewidth]{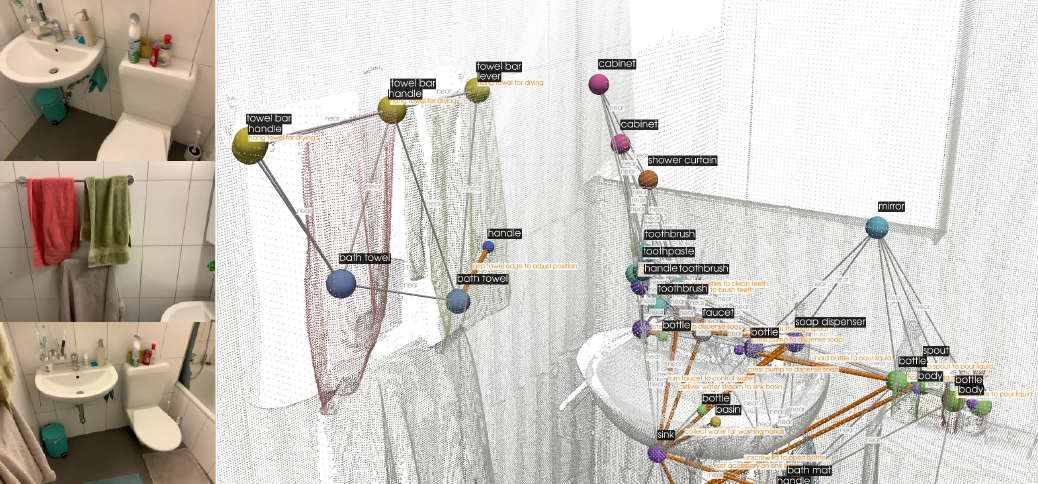}%{figures/qual_8bathroom_1.pdf}
    \caption{Unified 3DSG of the bathroom with a sink, soap, and toothbrushes}
    \label{sup:qual_5}
\end{subfigure}
\begin{subfigure}[t]{\linewidth}
    \centering
    \includegraphics[width=\linewidth]{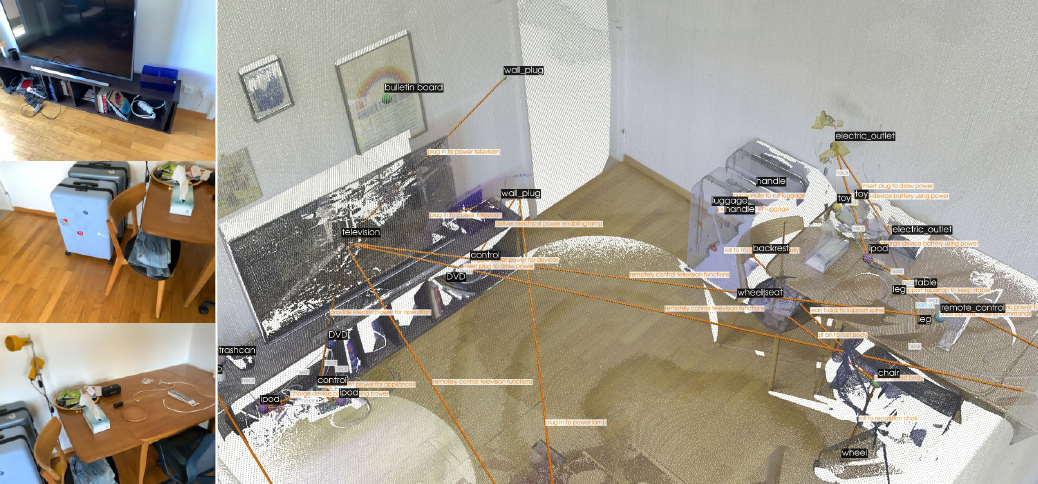}%{figures/qual_2livingroom_1.pdf}
    \caption{Unified 3DSG of the living room with a TV, a remote control, and luggage}
    \label{sup:qual_6}
\end{subfigure}
\caption{\textbf{Qualitative results of the OP3DSG.}}
\label{sup:qual}
\end{figure}
\clearpage

\subsection{Qualitative Results}
\label{appendix:qual}

\Cref{sup:qual} visualizes the unified 3DSGs generated in diverse indoor environments. The proposed framework successfully recognizes not only large objects (\eg, table, chair, bed, and cabinet), but also small objects (\eg, book, pen, and spatula) and fine-grained parts (\eg, knob, handle, and leg). In particular, \Cref{sup:qual_3} shows that multiple knobs are correctly associated with their corresponding cabinets through \textit{part of} relations, while also being assigned appropriate functional affordances such as \textit{pull to open}. In \Cref{sup:qual_4}, home appliances such as a toaster, coffee machine, and pot are correctly localized and labeled, and the generated graph further captures their functional dependency on a nearby electrical outlet through the \textit{provide power} relation. In \Cref{sup:qual_5}, the proposed method successfully reconstructs three distinct toothbrush nodes despite their thin and small geometry, showing accurate correspondence to the real environment. The results suggest that the proposed part-level 2D detection and 3D fusion pipeline can effectively support fine-grained node grounding and relation reasoning.

Nevertheless, several failure cases remain due to limitations in perception. For example, in \Cref{sup:qual_5}, a single towel bar is fragmented into three separate nodes because it is partially occluded by towels and appears as disconnected segments. In the same scene, the reflected shower curtain in the mirror is mistakenly recognized as a real object, leading to a false positive node. In \Cref{sup:qual_6}, several thin items are misrecognized as remote controls, resulting in multiple spurious objects that are further connected to the TV. These examples indicate that the remaining errors mainly arise from perceptual ambiguity caused by occlusion, reflection, and thin-object misclassification.

\section{Implementation Details} 
\phantomsection
\label{appendix:Implement}

\subsection{Framework}
\label{appendix:impleModel}

We employ RAM~\cite{zhang2024recognize} for image tagging and VLPart~\cite{sun2023going}, which is pretrained on the LVIS~\cite{gupta2019lvis} and PACO~\cite{ramanathan2023paco} datasets, for open-vocabulary detection. We further utilize SAM~\cite{kirillov2023segment} to generate instance masks from each detected region. For LLM-based reasoning, we use GPT-5~\cite{singh2025openai} (\textsc{gpt-5-2025-08-07}). In our experiments, we set $\tau_S=0.1$ and $\tau_V=0.9$ for visual part feature extraction. During the 3D fusion stage, the weighting coefficients $w^c_{1}$, $w^c_{2}$, and $w^c_{3}$ are set to 0.2, 0.4, and 0.4, respectively, while $w^p$ is set to 0.4.
% We also construct an object--part knowledge base manually, which contains 61 unique part categories associated with 88 object categories, resulting in 191 object--part pairs. 

% \subsection{LLM Prompts}
% \label{appendix:implePrompt}
% In our framework, we employ three types of LLM agents responsible for local relation reasoning, remote relation reasoning, and affordance reasoning. The corresponding prompts for each agent are presented in \Cref{sup:prompt}. Each prompt is structured into three main sections: \textit{identity}, \textit{workflow steps}, and \textit{rules}. The identity section defines the role and reasoning scope of the LLM agent. The workflow steps describe a structured reasoning procedure that the model should follow when inferring relations from the input graph. Finally, the rules section specifies constraints and output guidelines to ensure consistent graph construction and to reduce hallucinated relations during generation.

% \input{ECCV/section/prompt}

\subsection{Real-World Applications}
\label{appendix:app}
We demonstrate that the generated 3D scene graph by OP3DSG during indoor exploration can serve as a structured input representation for LLM-driven downstream tasks. In all experiments, we employ the GPT-5~\cite{singh2025openai} model for high-level reasoning. For Question Answering and Text Query tasks, the LLM receives the generated 3DSG together with instructions that require identifying nodes in the graph corresponding to the user’s query. Based on these instructions, the LLM performs reasoning over the structured graph to generate answers or determine the target node for navigation. For the Task Planning task, the LLM takes a high-level user goal as input and decomposes it into a sequence of executable commands at the part level. To execute the resulting action sequence on the robot, we utilize an open-source action executor provided by HomeRobot~\cite{Yenamandra2023HomeRobotOM}. For navigation, the generated 3DSG is used together with the reconstructed 3D map of the environment. A deterministic planner based on the A* algorithm~\cite{hart1968formal} computes the navigation path to the target location.

\begin{table}[t]
\caption{\textbf{Refined relation labels in UniGraph3D.}}
\label{sup:edge_refinement}
\centering
% \tablesize
\tinysize
\begin{tabular}{p{0.45\linewidth} p{0.53\linewidth}}
\toprule
\multicolumn{1}{c}{\textbf{SceneFun3D$^\dagger$~\cite{delitzas2024scenefun3d, zhang2025open}}} & \multicolumn{1}{c}{\textbf{FunGraph3D~\cite{zhang2025open}}} \\
\specialrule{0.06em}{2pt}{-2pt}
\begin{itemize}[leftmargin=8pt, itemsep=0pt, topsep=0pt, label=\raisebox{0.25ex}{\tiny$\bullet$}]
    \item press to open or close, or adjust the setting
    \item rotate to control the water flow
    \item press or rotate to open
    \item push to flush
    \item pull or rotate to open or close
    \item rotate to adjust the temperature
    \item rotate or press to adjust the setting
    \item rotate to adjust the setting
    \item rotate to open or close
    \item pull to open or close
    \item control, turn on or turn off
    \item press or rotate to flush
    \item press or rotate to control the water flow
    \item rotate to adjust setting or temperature
    \item pull to open or close a drawer
    \item rotate to adjust the setting or open or close
    \item control the water flow
    \item provide power
    \item press or rotate to control the water flow
    \item control
    \item rotate to flush
\end{itemize}
&
\begin{itemize}[leftmargin=13pt, itemsep=0pt, topsep=0pt, label=\raisebox{0.25ex}{\tiny$\bullet$}]
    \item press or rotate to control the drain water flow
    \item control to drain the water flow
    \item rotate to adjust the temperature
    \item press to flush
    \item pull to open or close a cabinet or drawer
    \item control
    \item rotate to adjust the setting or temperature, or to open or close
    \item press to open or close, or to adjust the wind flow
    \item rotate to open or close
    \item press or rotate to control the water flow
    \item rotate to open or close, adjust the setting
    \item provide power
    \item rotate to adjust the setting
    \item rotate to open or close, or to adjust the setting and temperature
    \item rotate to open or close, or to adjust the temperature and setting
    \item control, turn on or turn off
    \item pull to open or close
    \item operate or adjust the flow or setting
    \item press to start and operate
    \item push or press to open
    \item pull to open or close a drawer
    \item control the water flow
    \item rotate to adjust temperature or setting, open or close
    \item rotate to open or close, or to adjust the setting, time and temperature
    \item press to open or close, or to adjust the setting
\end{itemize}
\\
\specialrule{0.08em}{2pt}{2pt}
\multicolumn{2}{c}{\textbf{UniGraph3D}} \\
\specialrule{0.06em}{2pt}{-2pt}
\multicolumn{2}{p{0.98\linewidth}}{
\begin{itemize}[leftmargin=8pt, itemsep=0pt, topsep=0pt, label=\raisebox{0.25ex}{\tiny$\bullet$}]
    \item pull to open or close
    \item control the water flow
    \item rotate to adjust the setting or temperature
    \item press to start and operate
    \item turn on or turn off
    \item remote to control
    \item press to flush
    \item provide power
    \item push or press to open
    \item operate or adjust the flow or setting
    \item rotate to open or close
\end{itemize}
} \\
\bottomrule
\end{tabular}
\end{table}

\begin{figure}[t]
\centering
\begin{subfigure}[t]{0.53\linewidth}
    \centering
    \includegraphics[width=\linewidth]{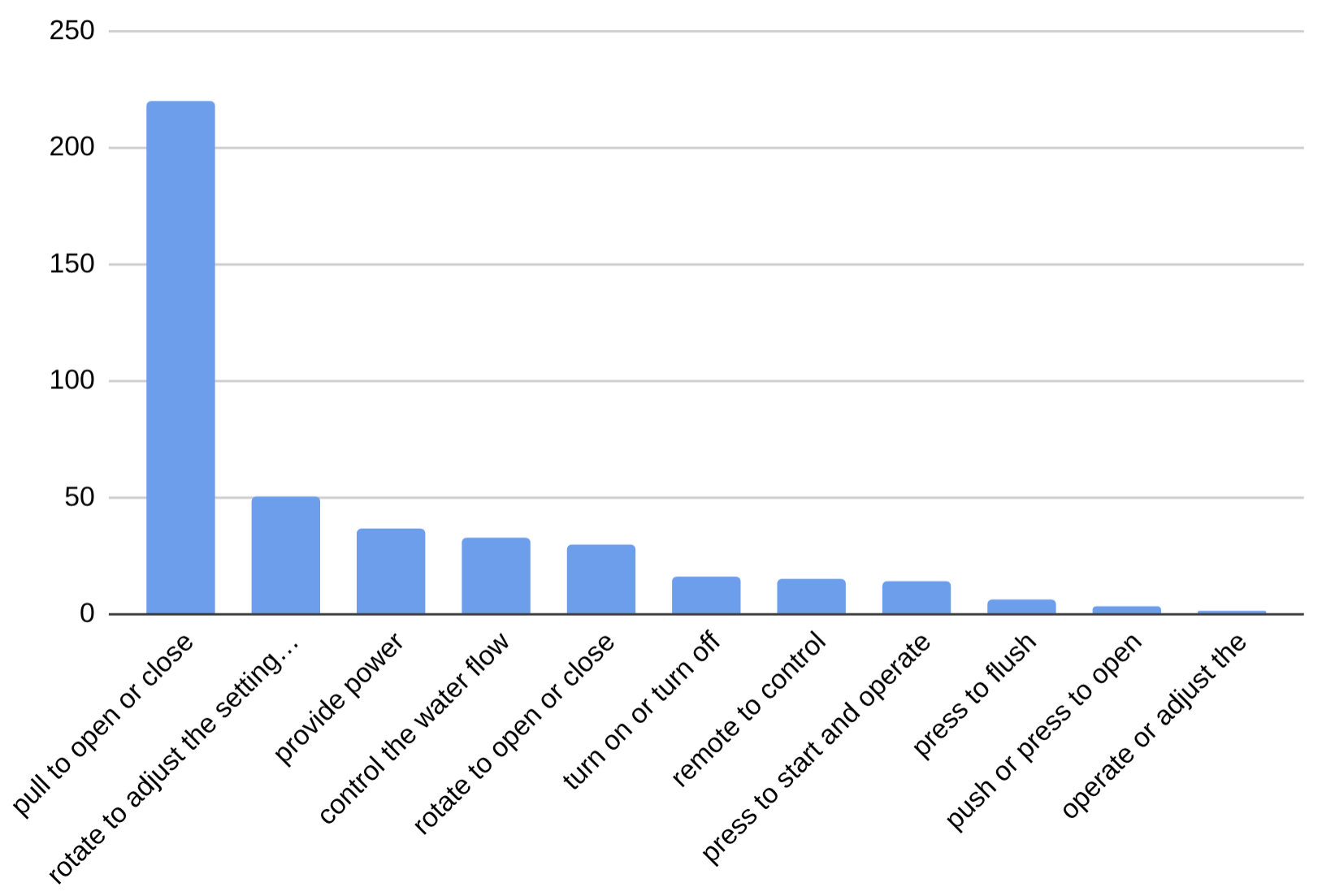}
    \caption{Functional relation}
    \label{sup:stat_a}
\end{subfigure}
\hfill
\begin{subfigure}[t]{0.44\linewidth}
    \centering
    \includegraphics[width=\linewidth]{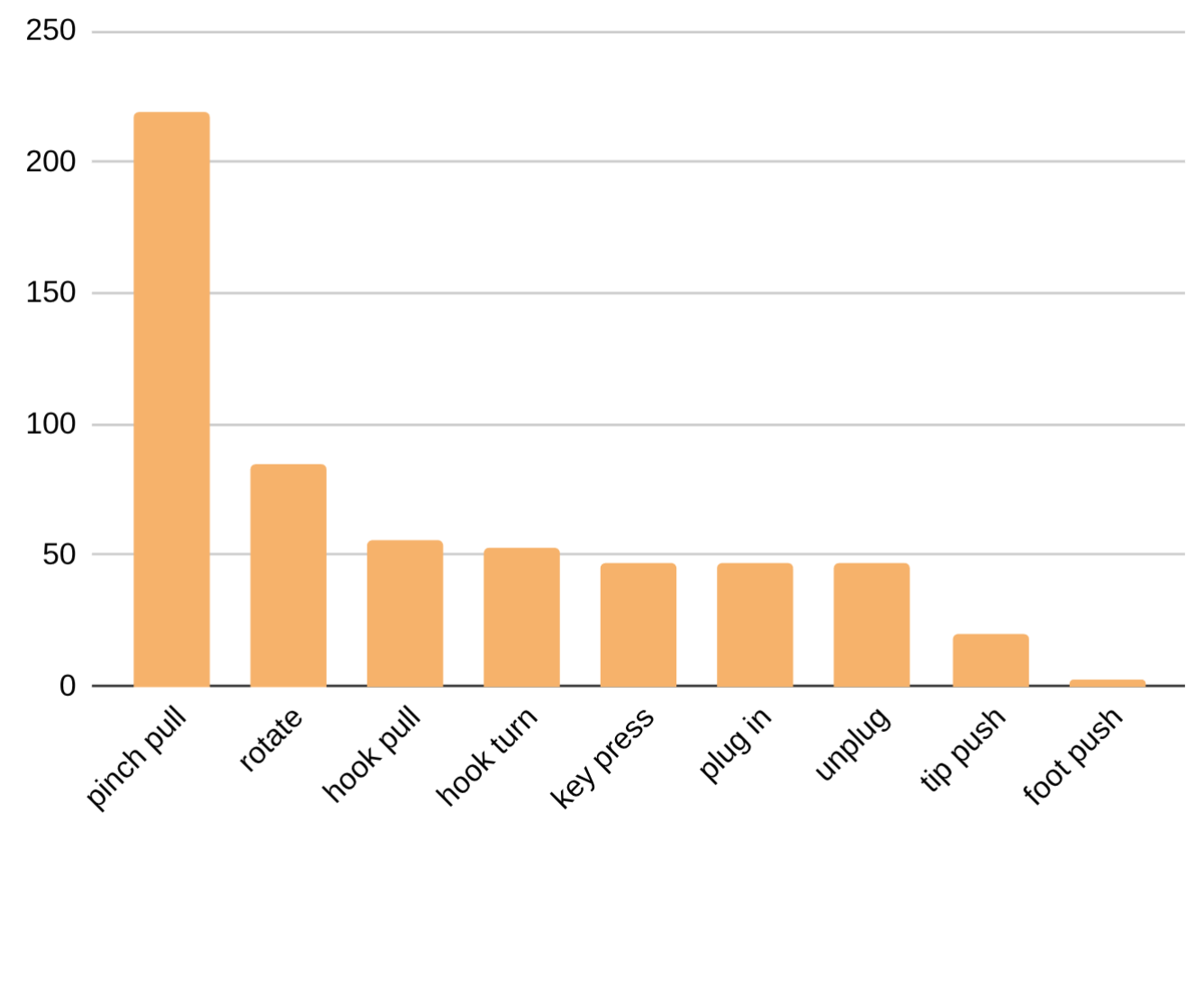}
    \caption{Affordance}
    \label{sup:stat_b}
\end{subfigure}
\caption{\textbf{The occurrence frequencies of functional relations and affordances.}}
\label{sup:stat}
\end{figure}

\section{Dataset Details}
\phantomsection
\label{appendix:Dataset}

\subsection{Relation Label Refinement}
\label{appendix:refine}
For our experiments, we leverage existing functional 3DSG datasets. However, the 21 relation labels in SceneFun3D$^\dagger$~\cite{delitzas2024scenefun3d, zhang2025open} and the 25 relation labels in FunGraph3D~\cite{zhang2025open} contain several ambiguous or semantically redundant categories. Such inconsistencies can undermine fair evaluation, particularly under a top-$k$ retrieval protocol, where semantically equivalent relations may be treated as distinct labels. As illustrated in \Cref{sup:edge_refinement}, identical or nearly identical phrases may appear as separate labels, such as \textit{``control''} and \textit{``control the water flow''}. In addition, several labels are assigned using subjective or ambiguous criteria, for example \textit{``rotate to adjust the setting''} and \textit{``rotate to adjust the temperature''}. Although these expressions often describe similar physical interactions, they are treated as different classes during evaluation, which can introduce unnecessary bias in performance measurement.

To address this issue, we analyze the relation labels across both datasets and consolidate semantically overlapping or ambiguous expressions into a set of 11 canonical relation labels. This normalization enables consistent evaluation across datasets and provides a clearer semantic space for functional relations. Since LLMs may generate different linguistic expressions for the same underlying relation, label normalization is particularly important for LLM-based approaches, as it ensures that semantically equivalent outputs are evaluated consistently. The occurrence frequencies of the functional relations are shown in \Cref{sup:stat_a}.

\subsection{Affordance Annotation}
\label{appendix:annot}
To maintain consistency with the existing dataset~\cite{delitzas2024scenefun3d}, we adopt the same 9 affordance labels and assign one or more to each node. The occurrence frequency of affordances is shown in \Cref{sup:stat_b}.

\begin{itemize}[leftmargin=13pt, itemsep=0pt, topsep=0pt, label=\raisebox{0.25ex}{\tiny$\bullet$}]
    \item \textit{rotate}: actions that adjust or control functionality through rotational manipulation of a knob or dial, \eg, thermostat knob, stove burner dial.
    \item \textit{key press}: interfaces composed of multiple pressable keys used to trigger discrete commands, \eg, remote control keypad, keyboard.
    \item \textit{tip push}: actions triggered by pressing a surface with the fingertip, typically requiring a small pushing force, \eg, light switch, elevator button.
    \item \textit{hook pull}: pulling interactions performed by hooking the fingers onto a handle-like surface, \eg, refrigerator handle, cabinet handle.
    \item \textit{pinch pull}: pulling actions executed using a pinching grasp, typically applied to small protruding components, \eg, drawer knob, cabinet knob.
    \item \textit{hook turn}: rotational manipulation performed after hooking the fingers around a handle or lever, \eg, door handle, faucet lever.
    \item \textit{foot push}: actions performed by pushing a surface using the foot, \eg, trash can pedal, foot-operated door pedal.
    \item \textit{plug in}: interactions that establish an electrical connection by inserting a plug into a power socket, \eg, power outlet, charging port.
    \item \textit{unplug}: interactions that disconnect an electrical connection by removing a plug from its socket, \eg, power outlet, extension cord socket.
\end{itemize}

% \clearpage

% \bibliography{main}

\end{document}